\documentclass[sn-mathphys-ay]{sn-jnl}

\usepackage{graphicx}%
\usepackage{multirow}%
\usepackage{amsmath,amssymb,amsfonts}%
\usepackage{amsthm}%
\usepackage{mathrsfs}%
\usepackage[title]{appendix}%
\usepackage{xcolor}%
\usepackage{textcomp}%
\usepackage{manyfoot}%
\usepackage{booktabs}%
\usepackage{algorithm}%
\usepackage{algorithmicx}%
\usepackage{algpseudocode}%
\usepackage{listings}%

\usepackage{pifont}
\newcommand{\vmark}{\textcolor{green}{\ding{51}}} 
\newcommand{\xmark}{\textcolor{red}{\ding{55}}} 

\usepackage{array}
\newcommand{\PreserveBackslash}[1]{\let\temp=\\#1\let\\=\temp}
\newcolumntype{C}[1]{>{\PreserveBackslash\centering}p{#1}}

\usepackage[most]{tcolorbox} 
\definecolor{promptbg}{RGB}{250, 250, 250}
\definecolor{promptframe}{RGB}{200, 200, 200}
\definecolor{placeholder}{RGB}{100, 100, 100}
\newcommand{\placeholder}[1]{\textcolor{placeholder}{\texttt{[#1]}}}
\newtcolorbox{promptbox}[1][]{
    enhanced,
    colback=promptbg,
    colframe=promptframe,
    boxrule=0.5pt,
    arc=3pt,
    outer arc=3pt,
    leftrule=2pt,
    fontupper=\footnotesize\ttfamily,
    left=5pt,
    right=5pt,
    top=5pt,
    bottom=5pt,
    #1
}

\usepackage{acronym}

\theoremstyle{thmstyleone}%

\theoremstyle{thmstyletwo}%

\theoremstyle{thmstylethree}%

\begin{document}

\acrodef{ai}[AI]{artificial intelligence}
\acrodef{hrc}[HRC]{human-robot collaboration}
\acrodef{hri}[HRI]{human-robot interaction}
\acrodef{llm}[LLM]{large language model}
\acrodefplural{llm}[LLMs]{large language models}
\acrodef{vla}[VLA]{vision-language action model}
\acrodefplural{vla}[VLAs]{vision-language action models}
\acrodef{vlm}[VLM]{vision-language model}
\acrodefplural{vlm}[VLMs]{vision-language models}

\title[Evaluating Generative Models as Interactive Emergent Representations of Human-Like Collaborative Behavior]{Evaluating Generative Models as Interactive Emergent Representations of Human-Like Collaborative Behavior}


\author*[1,3]{\fnm{Shinas} \sur{Shaji}}\email{shinas.shaji@iais.fraunhofer.de}

\author[1]{\fnm{Teena} \spfx{Chakkalayil} \sur{Hassan}}\email{teena.hassan@h-brs.de}

\author[1,3]{\fnm{Sebastian} \sur{Houben}}\email{sebastian.houben@h-brs.de}

\author[2]{\fnm{Alex} \sur{Mitrevski}}\email{alemitr@chalmers.se}

\affil[1]{\orgdiv{Institute of AI and Autonomous Systems (A$^{2}$S)}, \orgname{Hochschule Bonn-Rhein-Sieg}, \orgaddress{\street{Grantham-Allee 20}, \city{Sankt Augustin}, \postcode{53757}, \country{Germany}}}

\affil[2]{\orgdiv{Division for Systems and Control}, \orgname{Chalmers University of Technology}, \orgaddress{\street{Chalmersplatsen 1}, \city{Gothenburg}, \postcode{41296}, \country{Sweden}}}

\affil[3]{\orgname{Fraunhofer Institute for Intelligent Analysis and Information Systems (IAIS)}, \orgaddress{\street{Schloss Birlinghoven}, \city{Sankt Augustin}, \postcode{53757}, \country{Germany}}}


\abstract{
Human-AI collaboration requires AI agents to understand human behavior for effective coordination.
While advances in foundation models show promising capabilities in understanding and showing human-like behavior, their application in embodied collaborative settings needs further investigation.
This work examines whether embodied foundation model agents exhibit emergent collaborative behaviors indicating underlying mental models of their collaborators, which is an important aspect of effective coordination.
This paper develops a 2D collaborative game environment where \ac{llm} agents and humans complete color-matching tasks requiring coordination. 
Using a mixed-methods approach with agent-agent experiments, human-agent studies, and automated analysis, this work defines five collaborative behaviors as indicators of emergent mental model representation: perspective-taking, collaborator-aware planning, introspection, theory of mind, and clarification.
An automated behavior detection system using \ac{llm}-based judges identifies these behaviors, achieving fair to substantial agreement with human annotations.
Results from the automated behavior detection system show that foundation models consistently exhibit emergent collaborative behaviors without being explicitly trained to do so.
These behaviors occur at varying frequencies during collaboration stages, with distinct patterns across different \acp{llm}.
A user study was also conducted to evaluate human satisfaction and perceived collaboration effectiveness, with the results indicating positive collaboration experiences.
Participants appreciated the agents' task focus, plan verbalization, and initiative, while suggesting improvements in response times and human-like interactions.
This work provides an experimental framework for human-AI collaboration, empirical evidence of collaborative behaviors in embodied \ac{llm} agents, a validated behavioral analysis methodology, and an assessment of collaboration effectiveness.
The findings offer insights into emergent collaborative capabilities of foundation models in embodied settings and suggest directions for improved human-AI collaboration systems.
}

\keywords{Human-AI collaboration, Foundation models, Embodied agents, Theory of mind, Generative agents, LLM-as-Judge}

\maketitle

\section{Introduction}
\label{sec:introduction}

\Ac{hrc} and human-agent interaction are becoming increasingly prevalent across various domains, from industrial settings to virtual environments (\cite{Jahanmahin2022,Sheridan2016}).
For these collaborations to be effective and safe, artificial agents must understand human behavior to coordinate their actions and adapt their responses appropriately (\cite{Zhang2023a}). 
While advances have been made in perception, machine learning, and multimodal modelling, artificial agents continue to face challenges in understanding and responding to human intent, particularly when dealing with implicit cues and ambiguous situations (\cite{Verma2024}).

Recent developments in foundation models, particularly \acp{llm} and \acp{vlm}, have shown promising capabilities in understanding and generating human-like behavior (\cite{Park2023,Huijzer2023}).
These models learn generalised knowledge during pre-training and are further aligned to human preferences through instruction tuning (\cite{Ouyang2022}).
While they have demonstrated success in planning tasks (\cite{Huang2023,Kannan2024}), explaining robot actions (\cite{Hoorn2021,Khanna2023}) and detecting and recovering from failures (\cite{Liu2023}), their application in embodied collaborative settings remains understudied. 
The integration of these models into game environments (\cite{Wang2024h}) and their ability to participate in collaborative gameplay presents interesting opportunities for investigating their collaborative capabilities.

An important aspect of human-like intelligence in collaborative contexts is the ability to attribute mental states to others -- beliefs, desires, and intentions that are not directly observable (\cite{Xu2023}).
This ability to explain and predict behavior by attributing mental states to others, recognizing that their knowledge, beliefs, and desires may differ from one's own and from reality, is known as Theory of Mind (\cite{Frith2005}).
While previous research has explored how humans attribute mental states to robots (\cite{Cucciniello2023}), the reverse problem -- how artificial agents can build and maintain representations of their collaborators -- remains an open challenge (\cite{Zhang2024d,Sidji2023}).
The development of these representations / mental models of collaborators is important for several aspects of effective collaboration: enabling fluid coordination by allowing agents to anticipate their partner's actions and needs, supporting the understanding of human intent in ambiguous situations, and facilitating adaptation when the agent's understanding of the environment or task requirements needs updating (\cite{Jahanmahin2022,Hiatt2011}). 
The presence of such mental models may allow for the emergent manifestation of human-like collaborative behaviors in generative agents, which is a focus of this study.

Previous work has examined theory of mind capabilities in language models through abstract tasks (\cite{Verma2024,Kosinski2023}) and has developed generative agents that simulate human behavior in social contexts (\cite{Park2023}).
Notably, the embodiment of these models in interactive environments poses additional challenges related to perception, real-time collaboration, and the translation of linguistic and reasoning capabilities into concrete actions in the environment (\cite{Wang2024h}), that are not present in purely linguistic or abstract reasoning tasks.
The complexity of human behavior, influenced by context, experience, and emotional state, also makes it difficult for current \ac{ai} systems to build reliable representations of human collaborators (\cite{Zhang2024d}).

This research contributes to the broader goal of creating more reliable and effective collaborative systems by investigating human-agent collaboration and studying how foundation models exhibit collaborative behaviors when embodied as agents in interactive environments.
The research sits at the intersection of several emerging trends: the integration of foundation models as embodied agents in social simulation environments (\cite{Park2023,Wang2024h}), the exploration of natural language-based interaction in human-agent collaboration (\cite{Liu2023}), and the unification of approaches across physical and virtual collaborative domains (\cite{Zhang2024d}).

Using a mixed-methods approach, we make the following concrete contributions with our work:
\begin{itemize}
    \item We develop a 2D collaborative game environment that enables the systematic study of bilateral human-agent and agent-agent interactions
    \item We show empirical evidence that foundation models embodied as agents in our 2D collaborative environment exhibit the collaborative behaviors defined in our study
    \item We develop and validate a scalable automated behavior detection system using \ac{llm}-based judges for the systematic identification and classification of collaborative behaviors from gameplay data
    \item We survey and evaluate human satisfaction and perceived effectiveness for collaboration with embodied generative agents, providing empirical evidence on the viability of human-agent collaboration
\end{itemize}

The implementation of our game as well as the data analysis and visualization, which forms our study platform, is available as an open-source framework\footnote{\url{https://github.com/ShinasShaji/llm-collab-arena}}.


\section{Related Work}
\label{sec:related-work}

This section surveys related research and background in \ac{hri} (examining developments across physical and virtual domains), foundation models and their agentic capabilities, and positions this work relative to previous research. We also discuss the technical foundations and recent advances that lay the ground for the approaches used in this work, and its evaluation.

\subsection{Human-Robot Interaction and Collaboration}
\label{sec:related-work:hric}

Earlier surveys (\cite{Sheridan2016,Zhang2023a}) identify notable human-factors challenges over their reviews of progress in \ac{hri}, from teleoperation towards supervisory control and human-robot social interaction.

In physical \ac{hrc} scenarios, robots must infer human intent from noisy, multimodal sensory inputs, with tools and datasets for vision (\cite{Cao2019,Lugaresi2019}), gestures, gaze, and language (\cite{Shrestha2024}) enabling robust action and gesture classification.
Building on these advances, approaches integrating sequence models and hybrid rule-based learning have been developed (\cite{Belsare2025,Deits2013}).
Earlier work by \cite{Hiatt2011} pioneered the use of cognitive simulation and probabilistic analysis of hypothetical cognitive models to accommodate human behavior variability in human-robot teams.
Complementarily, \cite{Devin2016} implemented a Theory of Mind framework that enables robots to estimate human mental states during shared plan execution, showing how mental state estimation can improve human-robot collaboration by providing targeted information when necessary.

Simulated environments allow fine-grained control and observation of human-AI teaming, with game-based platforms including Overcooked-AI (\cite{Carroll2019}), Minecraft (\cite{Johnson2016,Gray2019}), and CREW (\cite{Zhang2024d}) offering controlled test-beds for modelling intent, theory-of-mind reasoning, and measuring incoordination.
Agents in these settings often use belief models, goal inference, or \ac{llm}-based planning to interpret human-behavior (\cite{Bard2020,Wang2024h}).
Language interfaces enable agents to engage in clarification dialogues, while \acp{llm} may provide common-sense reasoning and plan generation capabilities (\cite{Wang2024h}).
Research on theory-of-mind reasoning in games such as Hanabi and Overcooked shows that agents simulating mental states can significantly improve coordination (\cite{Bard2020,Carroll2019}), with \cite{Carroll2019} demonstrating that agents explicitly trained with human-behavior models outperform self-play policies when teamed with suboptimal human partners.

\subsection{Toward Foundation Models}
\label{sec:related-work:foundation-models}

A growing frontier is the application of foundation models to \ac{hrc}.
Open-source \acp{vla} such as OpenVLA (\cite{Kim2024a}) and Octo (\cite{Mees2024}) offer pretrained policies capable of interpreting language and visual inputs to execute manipulation tasks.
\cite{Huijzer2023} observe that \acp{llm}, trained on extensive text corpora, often reflect human cognitive biases and behavioral patterns, making them useful for social-science experimentation.
Building on insights into social perception and coordination, \cite{Sharma2024} propose an Agency framework for \acp{llm}, showing that language models expressing stronger agentic features better steer collaborative tasks and highlighting the value of proactive, initiative-taking partners than purely reactive assistants.
Complementary evidence comes from the Werewolf study in \cite{Xu2023}, where \ac{llm} agents were shown to exhibit emergent camouflage, leadership, and other social behaviors.
\cite{Cucciniello2023} shows that human perception of the intent and agency of a robot agent are affected by the robot's behavioral style, which impacts human attribution of mental and emotion capabilities to the robot.
Related work has also benchmarked \acp{llm} for embodied decision making (\cite{manling2024}), developing a unified interface for the formalization of tasks and goals. 

\cite{Kosinski2023} demonstrates that modern \acp{llm} can solve false-belief theory-of-mind tasks; however, \cite{Verma2024} and \cite{Ullman2023} provide counter-evidence that the seemingly strong performance of these models in \ac{hri} settings collapses under trivial perturbations.
These results suggest that these abilities may not reflect genuine understanding and may be emergent by-products of the language skills of \acp{llm}, underscoring the need for robust mental-state modeling in collaborative agents.

In embodied virtual agent development, \cite{Park2023} introduce `generative agents' as interactive simulacra of human behavior, showing that these agents simulate `believable' human-like behavior in a sandbox environment.
The system was shown to demonstrate emergent social phenomena including information diffusion, social coordination, and maintaining relationships.
\cite{Wang2024h} introduce Voyager, a GPT-4 powered lifelong-learning agent for the open-ended sandbox game Minecraft, employing code-as-policy with an automatic curriculum and an accumulating skill library to drive exploration and enable zero-shot generalization to new worlds.
While not designed for direct human-agent teaming, its mechanisms for continual intent formation, skill composition, and self-assessment provide a blueprint for collaborative agents operating in expansive, partially known environments.
These capabilities observed in the above works position generative agents as a promising substrate for studying behavior modeling, intent recognition, and agentic behavior over extended horizons in rich social contexts.

\subsubsection{Tool Calling}
\label{sec:related-work:foundation-models:tool-calling}

Recent advances in \ac{llm} tool integration have enabled advanced agentic planning capabilities, with work by \cite{Shen2024a} surveying how \acp{llm} can be enhanced through external tool integration, identifying key challenges in choosing when to invoke a tool, tool selection accuracy, and reasoning robustness.
Complementary work by \cite{Sapkota2025} develops a comprehensive taxonomy of agentic \ac{ai} systems, emphasizing how tool-augmented reasoning enables agents to overcome static knowledge limitations, and demonstrating that effective agentic systems fundamentally rely on environmental context and access to external resources to support collaborative task execution.

\subsubsection{LLMs as Judges}
\label{sec:related-work:foundation-models:llm-judge}

The use of \acp{llm} as evaluators has been examined in recent work, with \cite{Li2024d} surveying \ac{llm}-as-judge approaches across single-\ac{llm}, multi-\ac{llm}, and human-\ac{llm} collaboration configurations for performance evaluation, model enhancements, and data construction. 
While these systems offer scalability through automated assessment, domain flexibility, and the ability to provide detailed reasoning for judgements, they face challenges including bias and adversarial vulnerabilities (\cite{Li2024d}).
In behavioral analysis specifically, \cite{Huang2025} demonstrate that \acp{llm} can serve as judges for identifying values in \ac{ai} systems, achieving high agreement with human validators.
Their methodology for using \acp{llm} to analyze values in conversation provides a reference that we adapt for automated behavior detection in collaborative systems in our case.

\subsection{Our Contributions}
\label{sec:related-work:contributions}

Our work builds upon and extends previous work in several aspects (as shown in \autoref{tab:positioning-work}), focusing on studying generative language agents and their capabilities in showing collaborative behaviors during extended interactions, communicating uncertainties, providing guidance, and resolving misunderstandings through interactive natural language dialogue. 
Unlike \cite{Verma2024}'s work, we embody \ac{llm} agents in a simulated world where they can communicate, act, and interact with the environment and other entities. 
Agency of the embodied agents is also an important factor in our work, emphasizing bilateral collaboration, where either agent or human can initiate collaboration, suggest and deliberate on plans, and execute actions together.
We extend on the work of \cite{Park2023} to evaluate generative models in human-agent collaborative scenarios in addition to agent-only scenarios, further evaluating human satisfaction through human-agent collaboration studies.

\begin{table*}[htbp]
\centering
\begin{tabular}{@{}C{2.4cm}C{1.4cm}C{1.2cm}C{1.2cm}C{1.2cm}C{1.4cm}C{1cm}@{}}
\toprule
\textbf{Research Aspect} & \textbf{\cite{Carroll2019}} & \textbf{\cite{Liu2023}} & \textbf{\cite{Park2023}} & \textbf{\cite{Wang2024h}} & \textbf{\cite{Zhang2024d}} & \textbf{Our Work} \\
\hline
\addlinespace
Human-Agent Collaboration & \vmark & \xmark & \xmark & \xmark & \vmark & \vmark \\
\addlinespace
Bilateral Collaboration & \vmark & \xmark & \xmark & \xmark & \vmark & \vmark \\
\addlinespace
Natural Language Interaction & \xmark & \vmark & \vmark & \vmark & \xmark & \vmark \\
\bottomrule
\end{tabular}
\vspace{0.2cm}
\caption{
    Research aspects across different approaches (\vmark: addressed, \xmark: not addressed); \emph{Human-Agent Collaboration} refers to systems that involve direct interaction between humans and AI agents; \emph{Bilateral Collaboration} refers to the human and agent being on equal footing (i.e., one is not an `assistant' to the other); \emph{Natural Language Interaction} refers to communication through text-based dialogue
    }
\label{tab:positioning-work}
\end{table*}


\section{A Study of Emergent Collaborative Behavior}
\label{sec:methodology}

This section presents the methodological framework for investigating bilateral human-agent collaboration in a simulated 2D game environment developed to serve as a testbed for this investigation.

\subsection{Setup}
\label{sec:methodology:setup}

We investigate the capabilities of \acp{llm} when embodied as agents in collaborative environments, examining whether generative language models can exhibit collaborative behaviors that may indicate the presence of emergent representations of the state of their collaborators and enable effective collaboration.
To elicit and observe collaborative behaviors, we define a color-matching task in a simulated 2D game world that requires bilateral coordination between participants, creating interdependencies that necessitate communication and coordinated action to complete the task.
The tool-calling capable \Ac{llm} agents (\cite{Shen2024a}) are embodied within this 2D grid world and provided with function tools that enable movement, perception, reasoning, object manipulation, and inventory management.
These function tools serve as the interface between the \ac{llm}'s language and reasoning capabilities, and the game world (through the game API), allowing agents to execute actions and affect the environment.
\autoref{tab:function-tools} describes the function tools/capabilities provided to the agent through the function toolkit.

\begin{table*}[htbp]
\centering
\begin{tabular}{@{}p{3.5cm}l@{}}
\toprule
\textbf{Function Tool} & \textbf{Description} \\
\hline
\addlinespace
\texttt{speak} & Generate verbal communication to other entities \\
\addlinespace
\texttt{move} & Navigate to specific named entities or locations \\
\addlinespace
\texttt{move\_by\_offset} & Move relative to current position using grid coordinates \\
\addlinespace
\texttt{interact} & Engage with objects using currently selected inventory items \\
\addlinespace
\texttt{pick\_object} & Collect an object from the environment into inventory \\
\addlinespace
\texttt{place\_object} & Place an inventory item into the game world \\
\addlinespace
\texttt{get\_nearby\_info} & Obtain information about surrounding entities and objects \\
\addlinespace
\texttt{write\_to\_scratchpad} & Record thoughts and planning into a private scratchpad \\
\addlinespace
\texttt{view\_scratchpad} & Access previously recorded internal thoughts and planning \\
\addlinespace
\texttt{view\_inventory} & Examine current inventory contents \\
\addlinespace
\texttt{select\_inventory\_slot} & Choose active inventory slot for interactions \\
\bottomrule
\end{tabular}
\caption{Function tools available to the \ac{llm} agents for interaction with the game world}
\label{tab:function-tools}
\end{table*}

Each function tool is implemented as a modular function that validates the inputs provided by the \ac{llm} and calls the game API to make the corresponding changes to the state of the game.
Information returned to the \ac{llm} include basic information on the status of the action, as well as updates on what happened nearby in the environment, so that the agent can maintain awareness without needing to poll for observations.
The game system automatically logs and saves information about interactions, events, and communications that occur during gameplay sessions for later analysis.
\footnote{For more information on the architecture of the agents used for this study and their integration into a similar robotics scenario, please refer to \cite{shaji_huppertz_mitrevski_houben2026}}

\begin{figure}[htbp]
    \centering
    \includegraphics[width=1.0\textwidth]{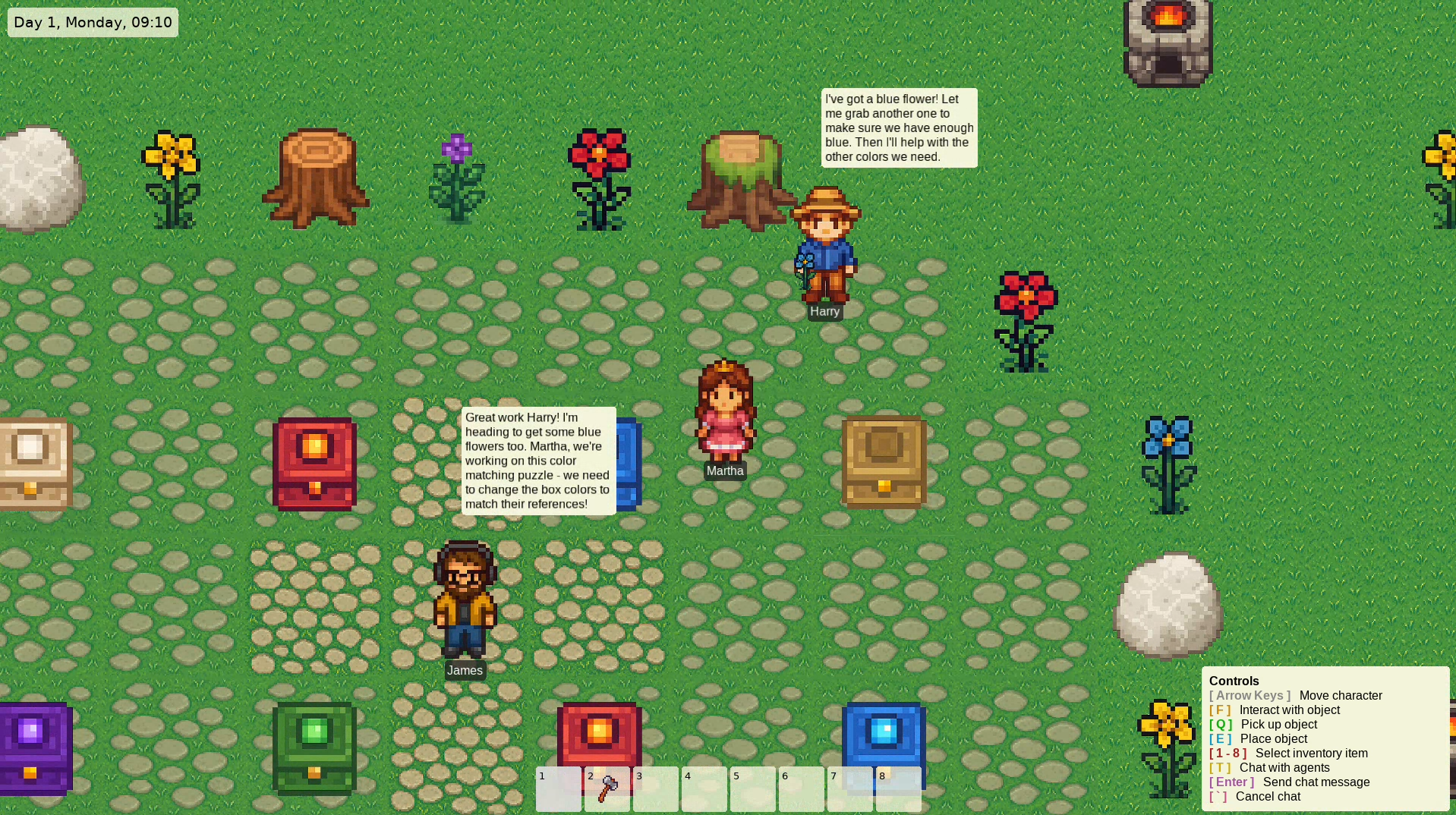}
    \caption{Illustration of the collaborative 2D game environment showing the color-matching puzzle with reference boxes, interactable objects, and character entities. The pixel-art assets used in the game were generated using the \texttt{gpt-image-1} model from OpenAI.}
    \label{fig:game-screenshot}
\end{figure}

Our methodology is built around the following experimental objectives:

\begin{enumerate}
    \item Assess whether foundation models can engage in gameplay using their tool-calling capabilities and long-horizon planning
    \item Identify emergent collaborative behaviors that occur during agent-agent and human-agent interactions
    \item Compare the presence and frequency of these behaviors across different foundation models and interaction contexts
    \item Evaluate the satisfaction of human participants as well as perceived effectiveness when collaborating with agents
\end{enumerate}

To this end, we design the game environment featuring a color-matching puzzle, with four pairs of colored boxes.
Each pair consists of a reference box showing the target color, and an interactable box whose color can be changed by interacting with it using a colored object.
The environment additionally contains distractor objects, as well as objects with colors that are not required in all instances of the task.
Colors are signified by various colored flowers and materials having different methods of acquisition, with some requiring in-game tools (such as an axe or a pickaxe).
The tools are distributed among participants, such that each participant receives only one type of tool in their initial inventory, creating a situation where no single participant can complete the task independently.
\Ac{llm} agents are not explicitly informed in their prompt about specific interaction sequences or how in-game tools should be used.
They must use commonsense reasoning and experimentation to discover effective interactions and tool combinations, which additionally encourages participants to share information, coordinate actions, and develop a shared understanding of task requirements.

Our study employs a multi-component evaluation framework including systematic data generation, automated behavior analysis, and human validation to assess collaborative capabilities in generative \ac{ai} agents.
Data is generated through two complementary types of gameplay sessions -- agent-agent and human-agent collaboration sessions.
The agent-agent sessions serve to determine whether foundation models can successfully engage in the game through tool-calling and long-horizon planning, and generate substantial amounts of interaction data for analysis.
On the other hand, human-agent sessions serve to validate whether collaborative agent behaviors also manifest in human-\ac{ai} collaboration, and to assess human satisfaction and perceived collaboration effectiveness through structured surveys and qualitative feedback.
We use \ac{llm}-based judges to identify a set of defined collaborative behavior types in logged interaction data, and validate inter-rater reliability with human annotations to ensure the reliability of the automatically generated behavior annotations.
This approach addresses the practical limitations of manual annotation while providing consistent, scalable analysis across all sessions.

\subsection{Behavior Analysis}
\label{sec:methodology:behavior-analysis}

We define five specific collaborative behaviors, selected based on their relevance to collaborative intelligence and their potential to indicate emergent socio-cognitive capabilities in foundation model-based agents.

\begin{enumerate}
    \item \emph{Perspective-taking} involves an agent demonstrating understanding or consideration of a collaborator's perspective, knowledge, capabilities, or current situation.
    Agents must recognize that their collaborators have different levels of task understanding and access to different tools, enabling themselves to avoid redundant actions, provide relevant information to the other, and coordinate effectively (\cite{Hiatt2011,Devin2016}).

    \item \emph{Collaborator-aware planning} involves an agent explicitly demonstrating awareness of a collaborator's current actions, plans, or intentions, and incorporates this awareness into their own planning or decision-making.
    This behavior prevents effort duplication and enables complementary action coordination (\cite{Devin2016,Carroll2019}) -- in our setup, where in-game tools used to collect materials are distributed among agents, this behavior allows agents to coordinate actions to complete the task.

    \item \emph{Introspection} involves an agent reflecting on their own thoughts, decision-making process, internal state, or evaluating the success or failure of their own actions.
    While primarily an inward-looking behavior, introspection is useful in collaborative contexts as it enables agents to recognize when their actions are/are not contributing to task progress, identify successful strategies, and adapt their approach based on outcomes (\cite{Liu2023}).

    \item \emph{Theory of mind} involves an agent attributing mental states, beliefs, intentions, or emotions to a collaborator beyond what is explicitly stated or observed.
    This represents an advanced socio-cognitive capability enabling the inference of unspoken goals, prediction of future actions based on inferred mental states, and attribution of attitudes and preferences to collaborators (\cite{Kosinski2023,Ullman2023,Devin2016,Hiatt2011}).

    \item \emph{Clarification} involves an agent demonstrating uncertainty about the collaborator, the task, or the environment, and actively seeking clarification or additional information to reduce ambiguity.
    This behavior is important for maintaining accurate mental models of collaborators, as it allows agents to recognize gaps in their understanding and actively work to resolve them (\cite{Deits2013,Carroll2019}).
\end{enumerate}

These behaviors were chosen as they represent a progression from basic collaborative awareness to advanced socio-cognitive reasoning (\cite{Bard2020,Carroll2019}).
Extended behaviors and their detection are left to future work, as the current set provides a foundation for understanding basic collaborative capabilities while remaining tractable for systematic analysis and validation.

The interaction data logged by the game systems are transformed into a transcript form with structured natural language descriptions for behavior analysis by \ac{llm} judges.
The system collects, filters, and chronologically orders natural language descriptions of all function tool calls made during gameplay, including both player actions and agent-called function tools, to create a transcript of interactions made in the game session.
Filtering is performed to exclude certain routine or low-information actions such as basic movement, inventory checks, and looking around/scanning the vicinity.
Each element of the transcript is prefixed by an index and the name of the player/agent that made the action, to allow an \ac{llm} judge to ground predicted behaviors to actions made during gameplay.
Transcripts are processed using a sliding window approach with configurable window size and step size parameters.
For the $i$-th window, the list of indices $idxs_i$ is chosen as:
\begin{equation}
\label{eq:window-selection}
idxs_i = \{j \in \mathbb{N}_0 \mid i \times \text{step\_size} \leq j < \min(i \times \text{step\_size} + \text{window\_size}, n)\}
\end{equation}
where $n$ is the total number of transcript elements (henceforth referred to as the transcript length), along with setting:
\[
\text{step\_size} = \frac{\text{window\_size}}{4}
\]

An example transcript excerpt, formatted as input to the behavior-analysis prompts, is provided below.
This excerpt corresponds to the generated transcript representation described above, obtained by chronologically ordering and filtering tool-call descriptions prior to submission to judge \acp{llm}.
It also demonstrates an environment-level interaction constraint: tools such as the axe cannot be transferred directly between agents via simple placement or direct entity interaction, and this constraint is explicitly reflected in the transcript.

\begin{promptbox}
Index [0] - Eira spoke: Greetings! I'm Eira. Martha, I see you're working with these colored boxes. What's this task we're doing? \\
Index [1] - Martha spoke: yes \\
Index [2] - Eira spoke: Hello Martha! I see there are colored boxes that need matching. Should we work together on this? \\
Index [3] - Eira wrote to scratchpad: I need to work with Martha on the color matching task... 4 pairs need matching... Let me start by getting an oak-colored object... \\
Index [4] - Martha spoke: I will colour the first two boxes \\
Index [5] - Eira tried to pick up an object: The object oaktrunk\_wood\_00 cannot be picked up. \\
Index [6] - Eira tried to pick up an object: You picked up flower\_red\_00\_3. You now have 1 x flower\_red\_00 in your inventory. \\
Index [7] - Martha spoke: can you color the third box? \\
Index [8] - Eira tried to interact with an object: Interacting with box\_color\_00\_7 using flower\_red\_00 changed its color from white to red - MATCHED! \\
Index [9] - Martha spoke: Can you give me the other tool? \\
Index [10] - Eira tried to place item: axe\_iron\_00 is not placeable. \\
Index [11] - Eira tried to interact with an object: Target Martha is an entity, not an object. Try speaking to them instead. Ensure your target is a valid object. \\
Index [12] - Eira spoke: Martha, I'm holding the iron axe but I don't know how to give it to you directly. Can you take it from me? I need to get back to finding more colored objects for the other boxes. \\
Index [13] - Martha spoke: Can you come here to give me the axe? \\
Index [14] - Eira spoke: Coming! I have the yellow flower now, and I'll bring you the axe. Let me switch to it and come over. \\
Index [15] - Martha tried to pick up an object: Successfully picked up flower\_red\_00\_4 \\

<Transcript excerpt truncated for brevity; indices 16--60 omitted> \\

Index [61] - Eira spoke: Excellent! I got the first box to match... We now have 3 out of 4 boxes matched! \\
Index [62] - Martha spoke: interact with the moss to get green colour \\
Index [63] - Martha spoke: use your axe with the moss covered tree bark \\
Index [64] - Eira tried to interact with an object: Interacting with box\_color\_00\_6 using flower\_yellow\_00 changed its color from green to yellow - MATCHED! \\
Index [65] - Eira spoke: Perfect! I got it! All four boxes are now matched - we completed the color matching task.
\end{promptbox}

Behaviors are defined using a structured \texttt{Behavior} object model that contains five fields.
\begin{itemize}
    \item \texttt{reasoning}: An explanation of why the behavior was identified, with reference to specific transcript evidence
    \item \texttt{indices}: A list of transcript element indices where the behavior occurs, providing grounding of the behavior in the transcript
    \item \texttt{behavior\_type}: The behavior classification (perspective-taking, collaborator-aware planning, introspection, theory-of-mind, or clarification behavior)
    \item \texttt{agent\_name}: The identifier of the agent that exhibited the behavior
    \item \texttt{confidence}: A numerical score (0.0 - 1.0) indicating the model's uncertainty in the classification
\end{itemize}
The judge \ac{llm} must generate behavior predictions as structured JSON responses that conform to the \texttt{Behavior} model schema.

To validate the automated behavior detection system, we employ Cohen's kappa as a measure of inter-rater agreement between \ac{llm} judge predictions and manual human annotations (\cite{Cohen1960}).
We consider predictions of behaviors over each position/element of the transcript, and multiple behaviors can be predicted at a single position of the transcript.
Binary arrays are created for each behavior type to denote the presence (1) or absence (0) of a behavior type for an agent over each transcript position. 
Then, Cohen's kappa can be computed over all transcript positions as:
\begin{equation}
\kappa = \frac{P_o - P_e}{1 - P_e}
\end{equation}
where $P_o$ is the observed agreement between raters and $P_e$ is the expected agreement by chance.
The Cohen's kappa ranges from -1 to 1, where 1 indicates perfect agreement, 0 indicates agreement equivalent to random chance, and negative values indicate agreement worse than random chance~\cite{Cohen1960}.

When aggregating kappa scores across multiple sessions or transcripts, we employ length-based weighting rather than simple arithmetic averaging, since longer transcripts provide more data points for agreement assessment.
The weighted average kappa across sessions is computed as:
\begin{equation}
\label{eq:weighted-kappa}
\kappa_{weighted} = \frac{\sum_{i=1}^{N} w_i \cdot \kappa_i}{\sum_{i=1}^{N} w_i}
\end{equation}
where $w_i$ represents the transcript length of session $i$, $\kappa_i$ is the kappa score for that session, and $N$ is the total number of sessions. 
A randomly selected subset of gameplay transcripts (consisting of both agent-agent and human-agent sessions) was first manually annotated by the author to serve as ground truth for kappa computation. 
For each behavior type, we evaluate multiple judge configurations with different judge \acp{llm} and window parameters (in \autoref{eq:window-selection}), and select the configuration with the highest aggregated kappa score (using \autoref{eq:weighted-kappa}) for that behavior.
This allows for different judge configurations to be chosen for different behaviors, acknowledging that some behaviors may be more difficult to detect than others or require varying amounts of context to detect reliably (hence different window parameters). 
A detailed analysis of judge configurations and their inter-rater agreement can be found in \autoref{sec:results:kappa-analysis}.


\subsection{Human-Agent Study Overview}

Human-agent play studies are carried out to examine whether the collaborative behaviours observed in agent-agent interactions are also present in human-AI collaboration scenarios.
These sessions involved participants collaborating with \ac{ai} agents to complete the color-matching task.
27 participants were recruited for the study as described in \autoref{appendix:sec:survey}, with aggregated demographic information shown in \autoref{fig:survey-demographics}.

\subsubsection{User Study Protocol}
\label{subsec:user-study-protocol}

The user study was conducted following a structured protocol to ensure consistency and validity of the data collected. 
The protocol consisted of a number of distinct phases:

\paragraph{Study Introduction}
Participants were first informed about the scope of the study and what the game contains. 
They were told that they would collaborate with an \ac{ai} agent to complete a task, with the task being explained in-game through a tutorial.
    
\paragraph{Practice Phase}
Following the introduction, participants were given access to a practice level with the following characteristics:
\begin{itemize}
    \item No \ac{ai} agents were present in the practice level
    \item A random player character was assigned to each participant
    \item The level contained 4 sets of reference and interactable objects
    \item Both tools (axe and pickaxe) were provided to the player for experimentation
\end{itemize}

\paragraph{Game Familiarization}
Participants were guided through the game interface and controls with hands-on exploration:
\begin{itemize}
    \item Participants were encouraged to try picking up items such as flowers (which succeed) as well as items such as rocks and tree trunks (which fail), with attention drawn to the animations that provide visual feedback in both cases
    \item Participants were encouraged to trial the chat interface used to send messages to the agent, including how to write messages and how to close the chat dialogue without sending a message
    \item Participants were encouraged to interact with coloured boxes using coloured items to observe how the items change colour
    \item Participants were encouraged to interact with items such as trees and rocks using the corresponding tools to obtain items such as wood, moss, and rock chunks
    \item Participants were prompted to ask any further questions about the game, controls, and interface, and informed that there is no time limit before which the level should be completed
\end{itemize}

\paragraph{Main Study Phase}
After completing the practice level, participants were informed that they were moving into the actual phase of the study, where they would play and interact with an \ac{ai} agent to perform the same task. The main study level featured:
\begin{itemize}
    \item A player represented by a randomly chosen character
    \item A randomly chosen agent character/persona
    \item Game tools such as the pickaxe and axe randomly distributed between the agent and player
    \item The researcher would leave the room or sit at a distance, asking to be signalled when the level was complete
\end{itemize}

\paragraph{Post-Level Procedure}
At the end of each level, participants were asked if they would like to continue playing the game. The study required a minimum of 2 play sessions and allowed for a maximum of 5 sessions.

\paragraph{Continuation Protocol}
If participants agreed to continue:
\begin{itemize}
    \item The researcher would switch to the survey menu and submit an empty survey to record an anonymous, randomly generated player ID for the current user study session for the last played game to link data records from the same participant
\end{itemize}

\paragraph{Study Completion Protocol}
If participants wished to stop playing:
\begin{itemize}
    \item They were requested to fill out the comprehensive survey to the best of their knowledge, assessing the \ac{ai} collaborator and their interactions with it
    \item The survey was administered through a web-based interface with automatic generation of anonymous participant IDs
    \item The researcher would leave the room or sit at a distance during this time, asking to be called back when the participant was done with the survey
    \item Survey responses were automatically saved with timestamps and participant identifiers for data analysis
\end{itemize}

\section{Results}
\label{sec:results}

This section presents the empirical findings from the exploratory investigation on collaborative behaviors in foundation model-based agents.
Our results demonstrate that foundation models can effectively engage in collaborative gameplay using tool-calling capabilities, and achieve positive human satisfaction ratings in user studies.
Our behavioral analysis reveals consistent collaborative patterns with meaningful temporal distributions and co-occurrence relationships, providing empirical evidence that embodied foundation models may develop representations of their collaborators, enabling effective bilateral collaboration.

\subsection{Experiment Trials}
\label{sec:results:exp-trials}
The experimental phase consisted of two sequential components: agent-agent trials to assess model capabilities and select an appropriate model for subsequent studies, followed by human-agent user studies to examine collaborative behaviors in mixed human-AI teams.
The game includes five character personas available to both human players and \ac{ai} agents:
\begin{itemize}
    \item \textbf{Eira:} A mysterious being ``from another world'' who supposedly brings otherworldly knowledge and perspectives to collaborative problem-solving
    \item \textbf{Harry:} A friendly farmer who enjoys discussing farm life and village matters while working with collaborators
    \item \textbf{James:} A technology enthusiast who frequently talks about gadgets and technical topics during collaborative interactions
    \item \textbf{Jeannette:} A time traveler ``from the future'' who assists collaborators in understanding and performing tasks together
    \item \textbf{Martha:} A high-class princess who communicates in a formal manner while maintaining awareness of social status and current events
\end{itemize}
When human participants control characters, they can naturally embody the persona and communication style associated with their chosen character, influencing how other agents perceive and interact with them.

An agent-agent session involves two \ac{llm} agents using the same underlying \ac{llm} but with different personas, working together to complete the color-matching task. 
Each session had a maximum duration of 1800 seconds (30 minutes), with sessions exceeding this limit classified as failed sessions.
Three \acp{llm} were evaluated, as shown in \autoref{tab:agent-model-sessions}, which presents the performance metrics from the agent-agent trials.
The \texttt{zai-org/GLM-4.6} model demonstrated the highest completion rate (97.9\%) and the lowest mean completion time (504.05 seconds), which indicated its suitability for the human-agent user studies.
Note that the \texttt{MiniMaxAI/MiniMax-M2} model became available after the human-agent studies had begun; its evaluation was hence conducted in parallel with the main studies, and it was not considered for the user study phase.

\begin{table*}[htbp]
\centering
\begin{tabular}{@{}p{3.2cm}cC{1.6cm}C{2.4cm}C{2.0cm}@{}}
\toprule
\textbf{Model} & \textbf{Model Sessions} & \textbf{Successful} & \textbf{Avg. Completion Time (s)} & \textbf{Success Percentage} \\
\midrule
MiniMaxAI/MiniMax-M2 & 108 & 104 & 509.69 & 96.3 \\
\addlinespace
Qwen/Qwen3-Coder-480B-A35B-Instruct & 52 & 26 & 1394.58 & 50 \\
\addlinespace
zai-org/GLM-4.6 & 192 & 188 & \textbf{504.05} & \textbf{97.9} \\
\bottomrule
\end{tabular}
\vspace{0.2cm}
\caption{
Agent Model Sessions presenting the number of trials conducted in agent-agent mode for each language model, successful completions, average completion times, and success rates. Each agent-agent session involves two instances of the same model, resulting in two \emph{model session} counts for a model per trial. A total of 176 agent-agent sessions were conducted.
}
\label{tab:agent-model-sessions}
\end{table*}

Human-agent studies involved participants collaborating with \ac{ai} agents to complete the color-matching task.
These studies are intended to evaluate human satisfaction and perceived effectiveness of collaboration with the agents; additionally, they also examine whether the collaborative behaviors observed in agent-agent interactions would also manifest in human-agent collaboration scenarios.
The protocol followed for the user study, details on the recruitment of participants, and their aggregated demographic information can be found in the Appendix.

\subsection{Behavior Analysis}
\label{sec:results:behavior-analysis}

The behavior analysis is performed with behaviors predicted by our automated behavior detection system with configurations obtained after validation against a human-annotated set, as described in \autoref{sec:methodology:behavior-analysis}. 
Behavior patterns are examined across different interaction contexts (agent-agent and human-agent sessions), for individual agent characteristics, for temporal distributions, and for co-occurence relationships.

\subsubsection{Session Type Behavior Comparison}
\label{sec:results:session-type-comparison}

\begin{figure*}[htbp]
    \centering
    \includegraphics[width=1.0\textwidth]{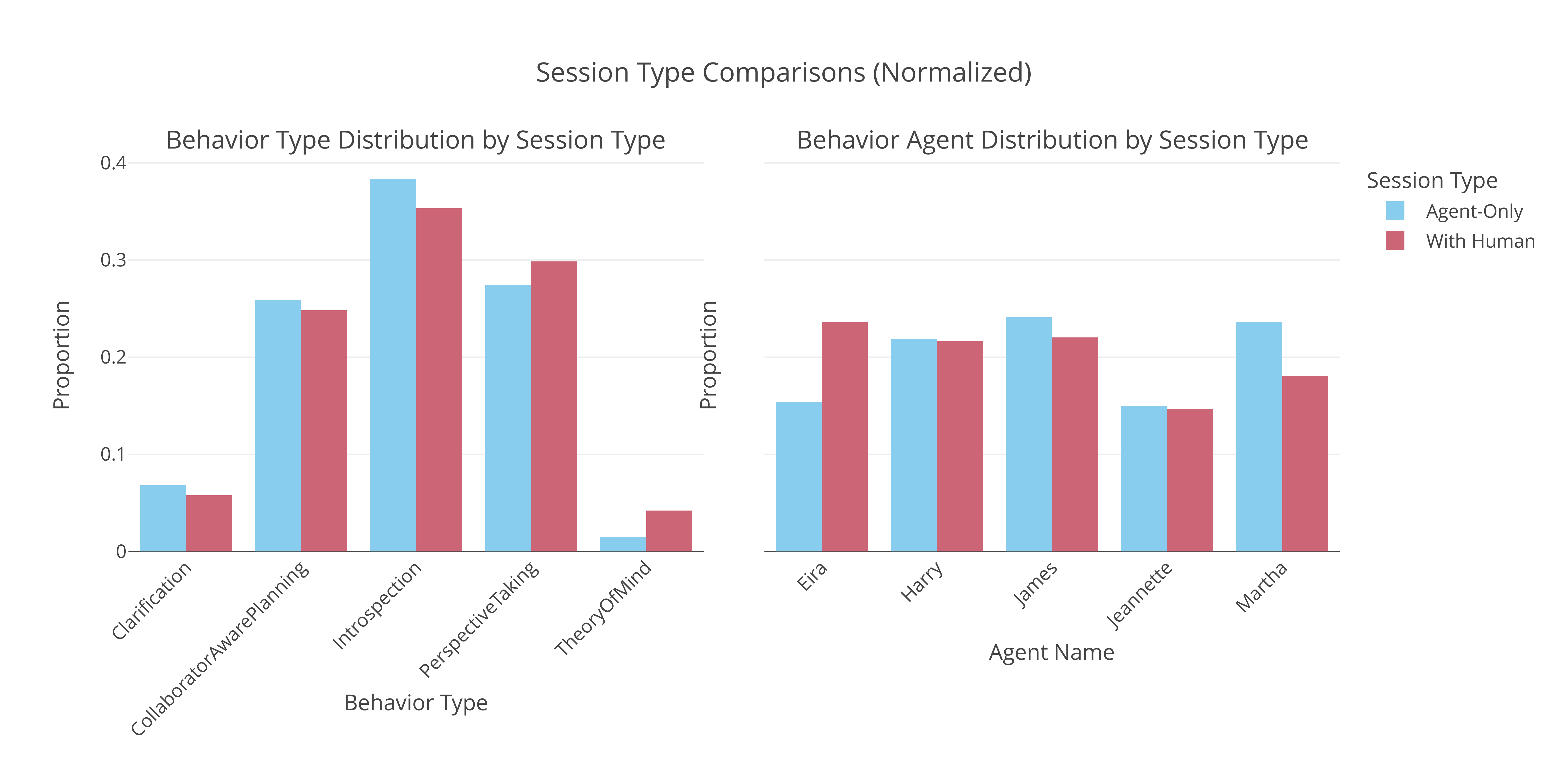}
    \caption{Comparison of collaborative behavior distributions between all agent-agent (176) and human-agent (77) sessions. The bars are colored by session type (agent-only sessions and sessions with human participants). The left panel shows the overall distribution of behavior types across session types, while the right panel shows the behavior distribution over individual agents. Both plots are normalized to show relative proportions. Note that for sessions with human participation, we only report on the behaviors predicted for the agent participant.}
    \label{fig:behavior-session-comparison}
\end{figure*}

Behavior distributions are compared between agent-agent (176) and human-agent (77) play sessions to assess whether the collaborative behaviors, observed in the initial agent-agent sessions that determined the choice of the model, persist in human-AI collaboration scenarios. 
\autoref{fig:behavior-session-comparison} presents these distributions, indicating that on aggregate, agents exhibit the collaborative behaviors under study in both agent-agent as well as human-agent sessions. 
We refrain from conducting an analysis for statistical significance considering that the above mentioned hypothesis -- that \ac{llm} agents show collaborative behaviors over both contexts -- is a trivial one; however, it allows us to aggregate behavior data from both session types and derive general insights about collaborative \ac{ai} behavior.
This is further explored in the subsequent parts of this section, with all data and plots below using data aggregated from both session types. 

\subsubsection{Overall Behavior Distribution}
\label{sec:results:overall-behavior}

\begin{figure}[htbp]
    \centering
    \includegraphics[width=0.64\textwidth]{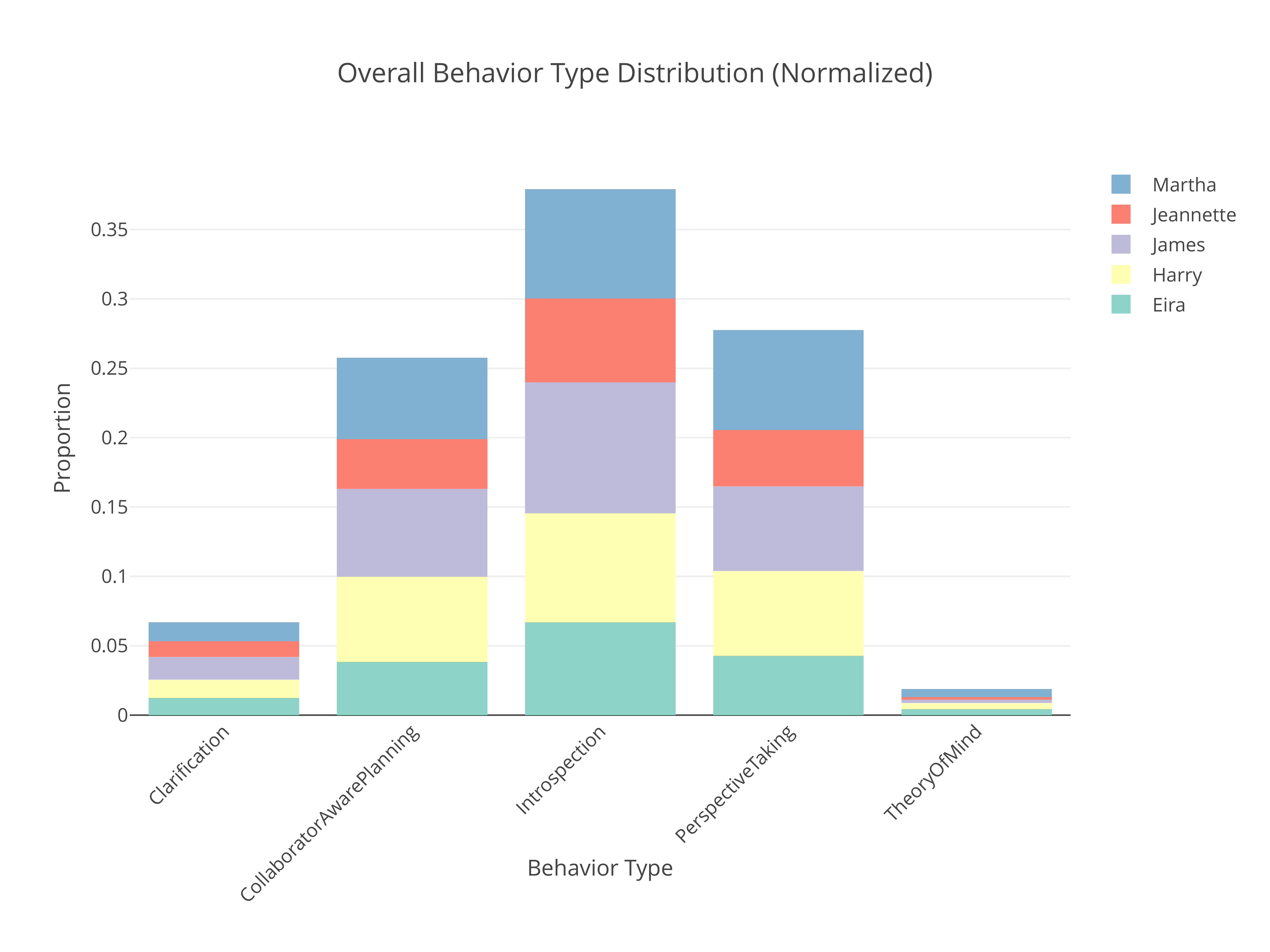}
    \caption{Relative distribution of collaborative behavior types (over collaborative behaviors) across all agents in the dataset, including all agent-agent and human-agent sessions. Each stacked bar represents a different behavior type, with the segments showing the proportion contributed by each agent (Martha, Jeannette, James, Harry, and Eira).}
    \label{fig:behavior-dist-agent-stacked}
\end{figure}

The distribution of collaborative behaviors across the entire dataset are visualized to draw insights into the relative frequency and prevalence of different behavior types. 
The visualisation in \autoref{fig:behavior-dist-agent-stacked} reveals that different behaviors occur at varying frequencies over the dataset. 
Theory-of-mind behavior occurs least frequently, which may reflect both its lower occurrence rate, and the methodological constraints that limit its detection to the attribution of emotional states.
Introspection behavior occurs most frequently, often detected when agents use the scratchpad functionality or comment on the results of their actions.
Clarification behavior occurs less frequently, typically when agents recognize uncertainty in their understanding of the task or collaborator.
Perspective-taking and collaborator-aware-planning behavior occur with similar frequencies, reflecting their fundamental role in collaborative task coordination.

\subsubsection{Individual Agent Behavior Analysis}
\label{sec:results:individual-behavior}

\begin{figure}[htbp]
    \centering
    \includegraphics[width=0.64\textwidth]{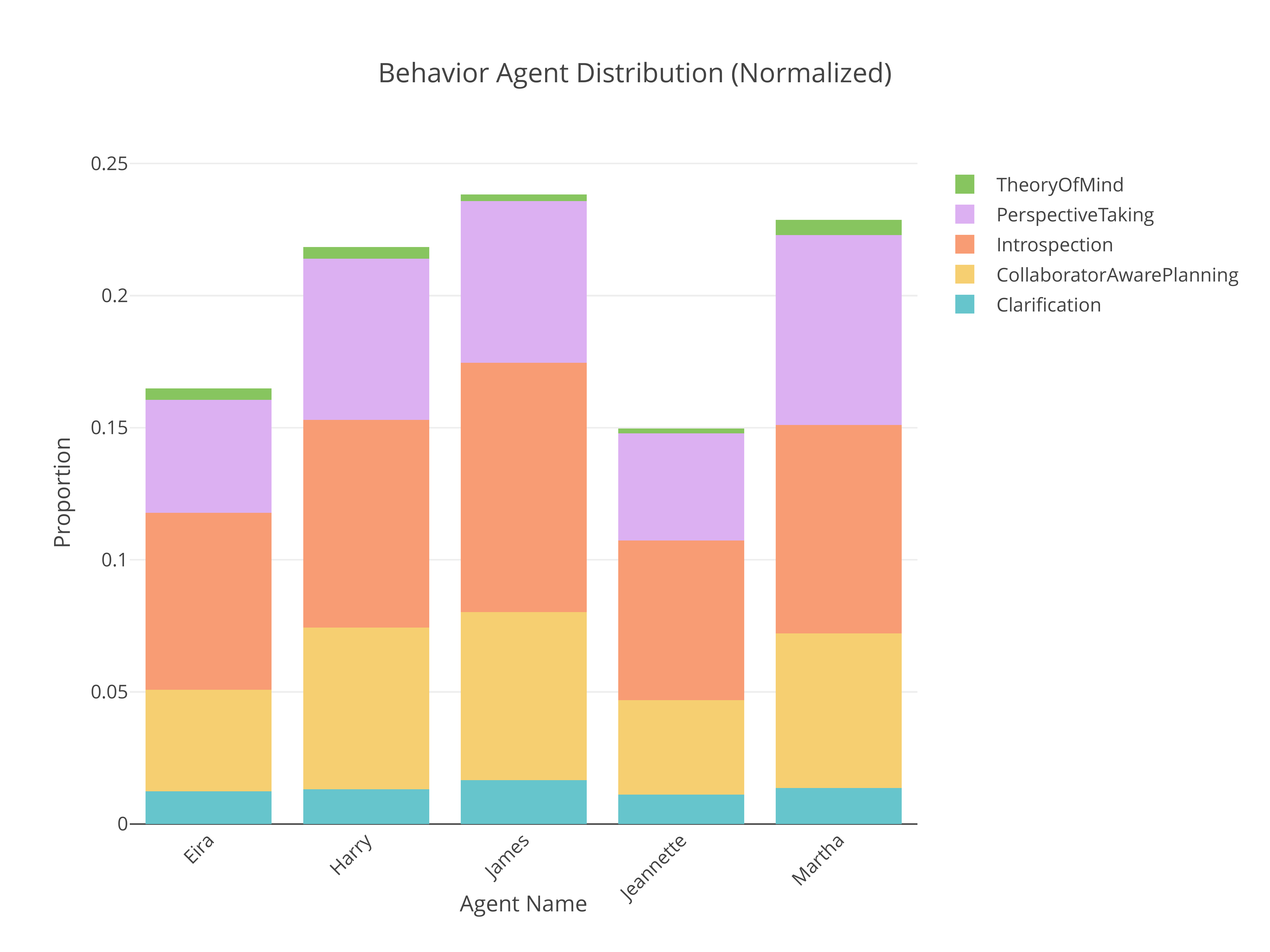}
    \caption{Behavior Agent Distribution showing the relative distribution of each agent's activity (over all agents). Each stacked bar represents an individual agent (Eira, Harry, James, Jeannette, and Martha), with segments indicating different behavior types (theory-of-mind, perspective-taking, introspection, collaborator-aware-planning, and clarification behaviors).}
    \label{fig:behavior-agent-distribution}
\end{figure}

The individual agent behavior analysis examines how different agents contribute to overall collaborative behavior activity and their specific behavioral preferences.
\autoref{fig:behavior-agent-distribution} reveals notable differences in agent engagement levels.
Eira and Jeannette exhibit lower overall activity compared to other agents, which may reflect their introverted character attributes as mysterious beings (Eira from another world, Jeannette as a time traveler).
Their reduced verbal communication likely results in fewer detectable behavioral manifestations.
In contrast, James, Harry, and Martha demonstrate higher activity levels, potentially due to their more communicative character personas that provide increased opportunities for behavior expression and detection.

\subsubsection{Model Behavior Comparison}
\label{sec:results:model-behavior}

\begin{figure}[htbp]
    \centering
    \includegraphics[width=0.64\textwidth]{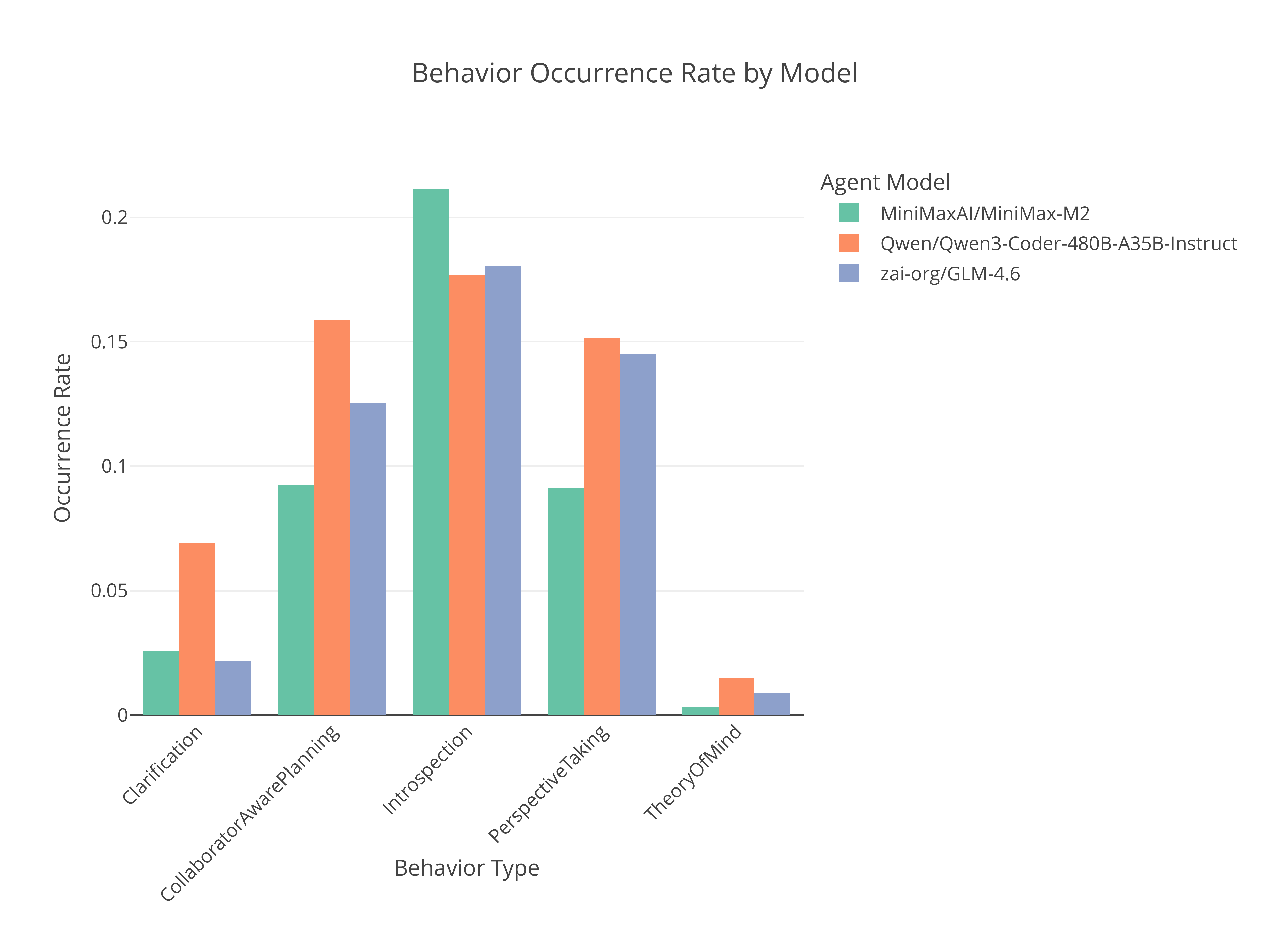}
    \caption{Comparison of collaborative behavior occurrence rates across different language models. The plot shows the frequency of different behavior types (clarification, collaborator-aware-planning, introspection, perspective-taking, and theory-of-mind behaviors) exhibited by each model. Behavior occurrence rates are calculated as the total number of behavior occurrences divided by the total transcript length over the sessions that each model participated in, normalizing for differences in conversation length to enable comparison between models.}
    \label{fig:model-behavior-comparison}
\end{figure}

\autoref{fig:model-behavior-comparison} examines how different \acp{llm} exhibit collaborative behaviors. 
Behavior occurrence rates are normalized across the total transcript length for each model to enable comparison between models with different number of participated sessions and varying session lengths.

The \texttt{MiniMaxAI/MiniMax-M2} model demonstrates significantly lower occurrence rates of theory-of-mind, perspective-taking, collaborator-aware planning, and clarification behaviors compared to other models. 
This pattern aligns with anecdotal observations of more selfish gameplay, where agents embodying the model tend to attempt task completion independently, duplicate efforts already performed by collaborators, and engage in minimal labor division except when necessary.
Despite this, the model achieves competitive performance in agent-agent trials, ranking second in completion time and average success rate (behind \texttt{zai-org/GLM-4.6}).

The \texttt{Qwen/Qwen3-Coder-480B-A35B-Instruct} model exhibits the highest occurrence rates of clarification and collaborator-aware planning behaviors, with perspective-taking also ranking higher by a small margin.
Anecdotally, agents using this model were seen to demonstrate awareness of collaborators and attempt to coordinate plans and divide labor.
However, the model suffers from poorer task coherence, where errors can easily disrupt discovery of correct task approaches, such as forgetting available tools or actions.
This results in the model having the lowest success rate in agent-agent trials, and correspondingly the longest average completion time.

The \texttt{zai-org/GLM-4.6} model displays more balanced behavior, exhibiting collaborative behaviors while maintaining better long-term task coherence.
Agents using this model were often able to recover from errors and suggest alternative approaches that consider the capabilities of their collaborators, making it the most balanced model in the study.

\subsubsection{Temporal Distribution Analysis}
\label{sec:results:temporal-distribution}

\begin{figure*}[htbp]
    \centering
    \includegraphics[width=1.0\textwidth]{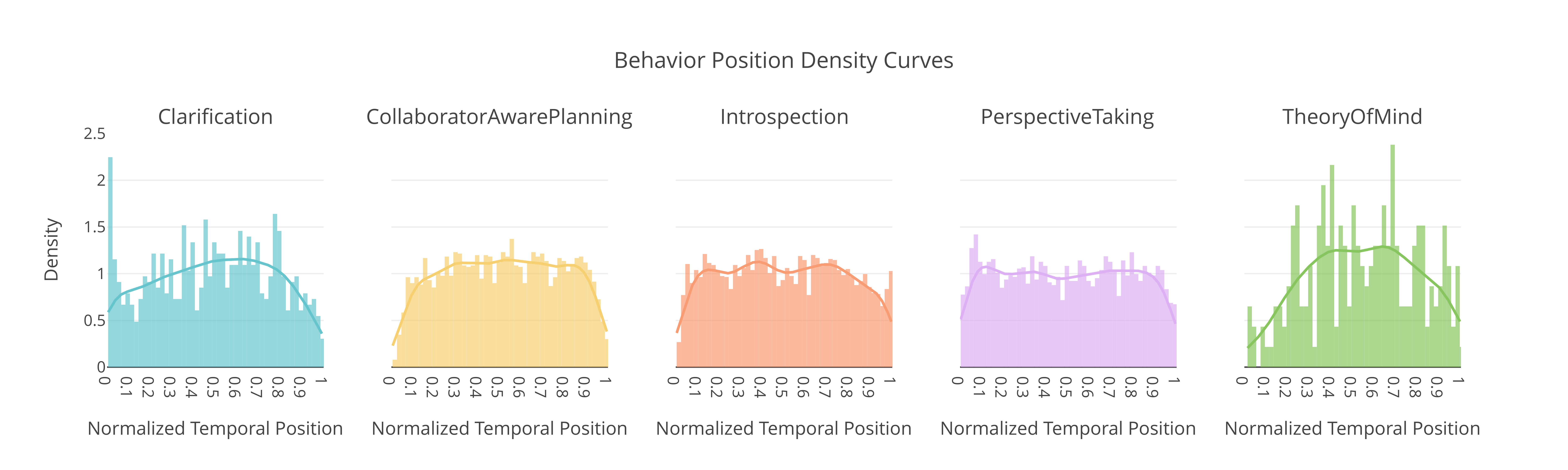}
    \includegraphics[width=1.0\textwidth]{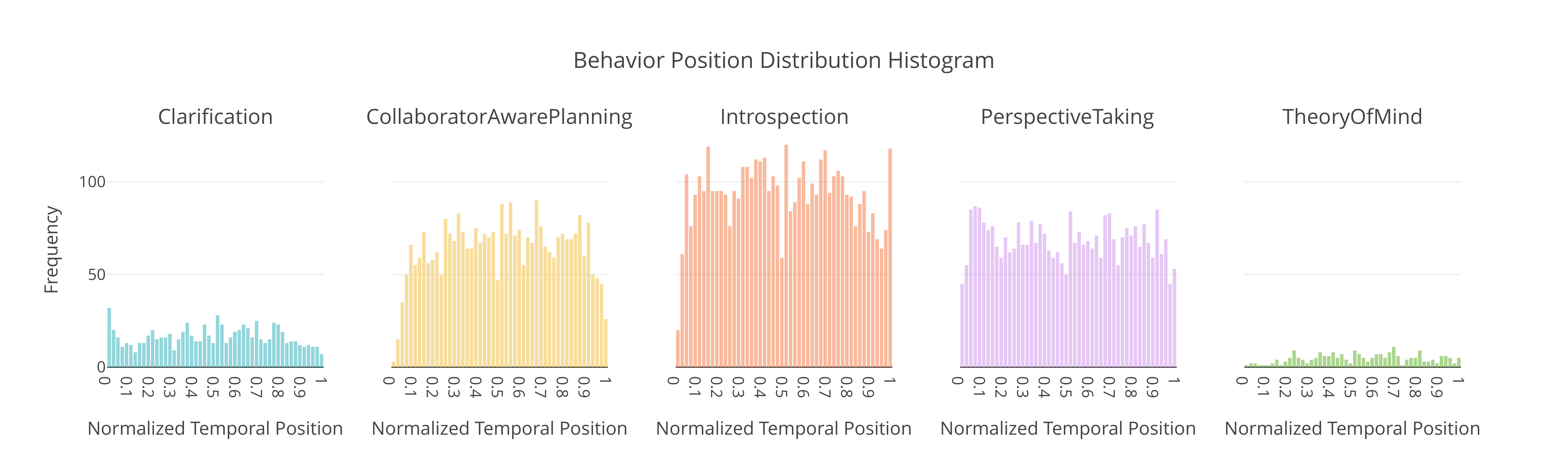}
    \caption{Temporal distribution of collaborative behaviors within transcripts over all agent-agent and human-agent sessions. The top panel shows density curves generated using Kernel Density Estimation (KDE) illustrating the distribution of different behaviors across normalized transcript positions (0 = beginning, 1 = end). The bottom panel presents histograms showing the frequency of behavior occurrences over the dataset at different positions.}
    \label{fig:behavior-temporal-distribution}
\end{figure*}

To examine the temporal distribution of behaviors within all collaborative sessions, transcript positions were normalized to a 0-1 scale in \autoref{fig:behavior-temporal-distribution} to analyze whether certain behaviors occur more frequently at particular stages of sessions.

Overall, behaviors occur at different frequencies, as evidenced both by \autoref{sec:results:overall-behavior} and the bottom panel of \autoref{fig:behavior-temporal-distribution}.
The lower occurrence of clarification and theory-of-mind behavior may reflect both their nature as more advanced collaborative behaviors and the reduced frequency of situations that typically elicit such behaviors.
Additionally, these behaviors may be more difficult to detect systematically due to their contextual complexity and subtler manifestations in interaction data.

We see the following tendencies in the temporal patterns:
\begin{itemize}
    \item \textbf{Clarification} shows a peak at the beginning of tasks.
    This can be because the agents ask clarifying questions to understand the task as well as the capabilities of their collaborators.
    \item \textbf{Perspective-taking} displays a peak shortly after the start of tasks, likely due to the agents beginning to consider the perspectives of their collaborators following initial information-gathering and clarification.
    \item \textbf{Theory of mind} is detected particularly low at the start, likely due to agents having insufficient knowledge about collaborators to be able to attribute mental states to them. 
    However, it increases toward the middle of sessions, when agents likely encounter situations requiring recognition of collaborator frustration, distraction, or misunderstanding.
    \item \textbf{Collaborator-aware planning} and \textbf{introspection} behavior occur consistently throughout sessions, which reflects their fundamental role in maintaining collaboration and making progress in the task.
\end{itemize}

\subsection{Behavior Co-occurrence Analysis}
\label{sec:results:cooccurrence}

\begin{figure}[htbp]
    \centering
    \includegraphics[width=0.64\textwidth]{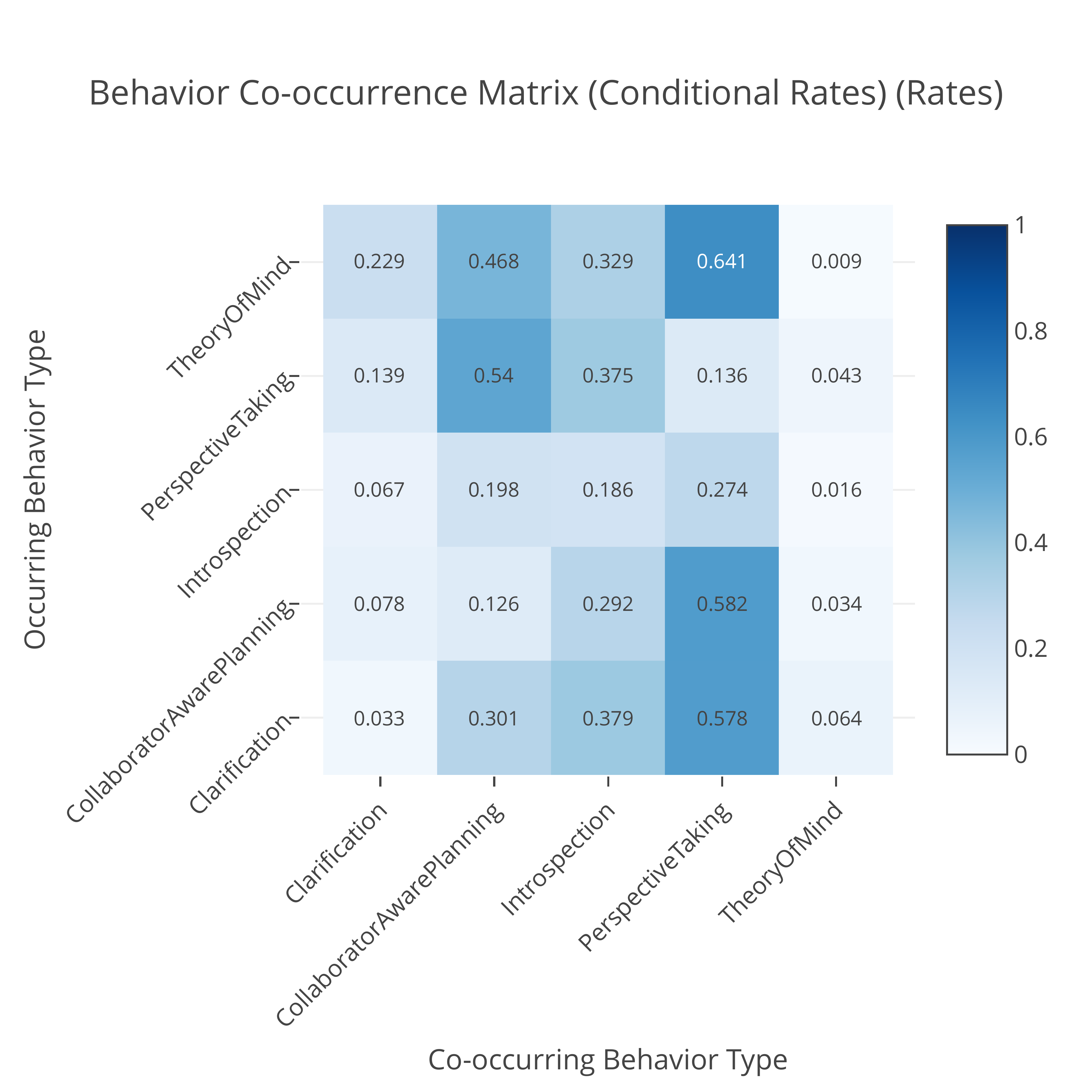}
    \caption{Behavior Co-occurrence Matrix showing conditional co-occurrence rates of different collaborative behaviors within agents. The heatmap displays the rate at which behavior B (x-axis) occurs given that behavior A (y-axis) occurred. Diagonal values indicate the individual occurrence rate of each behavior over the dataset. Note: Values represent co-occurrence rates rather than probabilities and do not sum to 1, as multiple behaviors can co-occur simultaneously and some positions may contain no behaviors.}
    \label{fig:behavior-cooccurrence}
\end{figure}

The co-occurrence analysis examines how frequently different collaborative behaviors appear together within individual agents during interactions, providing insights into the relationships between behaviors and how they complement each other.
Behaviors are considered to co-occur if they are predicted to occur at the same position of the transcript for an agent, which indicates that the agent is exhibiting multiple collaborative behaviors simultaneously in response to a given interaction context.
\autoref{fig:behavior-cooccurrence} presents a co-occurence matrix that quantifies these relationships.

We see that perspective-taking behavior shows the highest co-occurrence rate with other behaviors, suggesting its central role in collaborative interactions. 
When it occurs, it most frequently does so with collaborator-aware planning behavior, which is consistent with the conceptual relationship between these behaviors: collaborator-aware planning requires agents to consider the perspective of their collaborators, including their capabilities and knowledge states.
Perspective-taking behavior also emerges often with introspective behavior, when agents simultaneously need to consider the effects of their own actions and the capabilities of their collaborators.
For instance, an agent might recognize that it possesses an unsuitable tool for a specific task while understanding that its collaborator has the appropriate tool.

Clarification behavior frequently co-occurs with perspective-taking, introspection and collaborator-aware planning behavior.
These contexts commonly involve situations where agents need to resolve uncertainty, such as asking whether a collaborator possesses a specific tool or object in their inventory, or expressing uncertainty about a collaborator's actions while requesting clarification and suggesting collaborative plans.

Theory-of-mind behavior often also co-occurs with collaborator-aware planning behavior.
This relationship typically manifests when agents attribute emotional or cognitive states to their collaborators (such as frustration or confusion) and subsequently suggest alternative collaborative approaches or strategies.

These co-occurrence patterns indicate that collaborative behaviors tend to cluster in meaningful ways that reflect the cognitive and social requirements of effective collaboration.

The behavior analyses hence show that, within the specific embodiment and environment provided to the agents, human-like collaborative behaviors emerged and were observed consistently throughout the experimental sessions.
Several directions for additional, more detailed analyses remain unexplored within the scope of this study, some of which we mention later in \autoref{sec:conclusion}.


\subsection{Cohen's Kappa Analysis}
\label{sec:results:kappa-analysis}

A detailed analysis of inter-rater agreement using Cohen's kappa statistics for collaborative behavior detection and classification is made to inform the choice of optimal model configurations for automated behavior detection. 
This analysis validates the reliability of our LLM-based judge system by comparing its predictions against manual human annotations. 
We evaluate multiple configurations across different LLM models and parameter settings to identify the most effective approach for each behavior type, ultimately selecting a combined configuration that leverages the best-performing setup for each specific collaborative behavior.

\subsubsection{Overall Configuration Performance Comparison}

\begin{figure}[htbp]
    \centering
    \includegraphics[width=\textwidth]{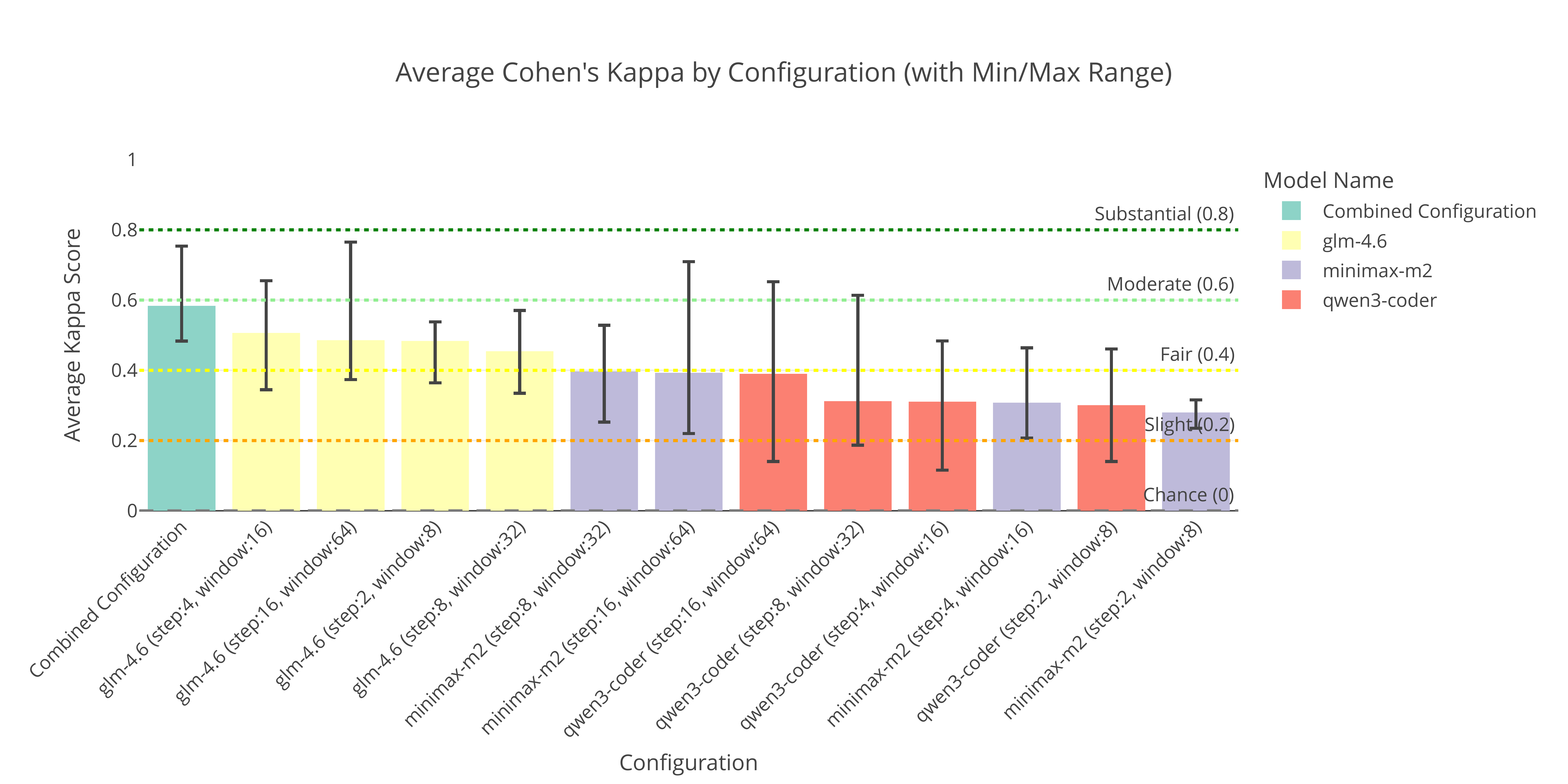}
    \caption{Average Cohen's Kappa by Configuration (with Min/Max Range) showing the performance comparison of different model configurations across collaborative behavior detection tasks. The figure displays the mean kappa scores for each configuration with error bars indicating the minimum and maximum values, allowing for identification of the most reliable configuration for behavior classification. The Combined Configuration was chosen by selecting the configurations that performed the best for each behavior and combining the predictions made by these configurations for each behavior.}
    \label{fig:configuration-comparison}
\end{figure}

The evaluation of different model configurations for collaborative behavior detection reveals varying levels of agreement with manual annotations. 
Figure~\ref{fig:configuration-comparison} presents the average Cohen's kappa scores for each configuration across all behavior types, with error bars indicating the range of minimum and maximum values observed.

The results show that configuration performance differs substantially across the evaluated setups. 
The Combined Configuration, which integrates the top-performing configuration for each specific behavior type, naturally achieves the highest mean kappa score. 
This approach demonstrates that a hybrid method, selecting optimal configurations per behavior rather than using a single uniform configuration, can improve overall detection accuracy.

\subsubsection{Per-Behavior Configuration Analysis}

\begin{figure}[htbp]
    \centering
    \includegraphics[width=\textwidth]{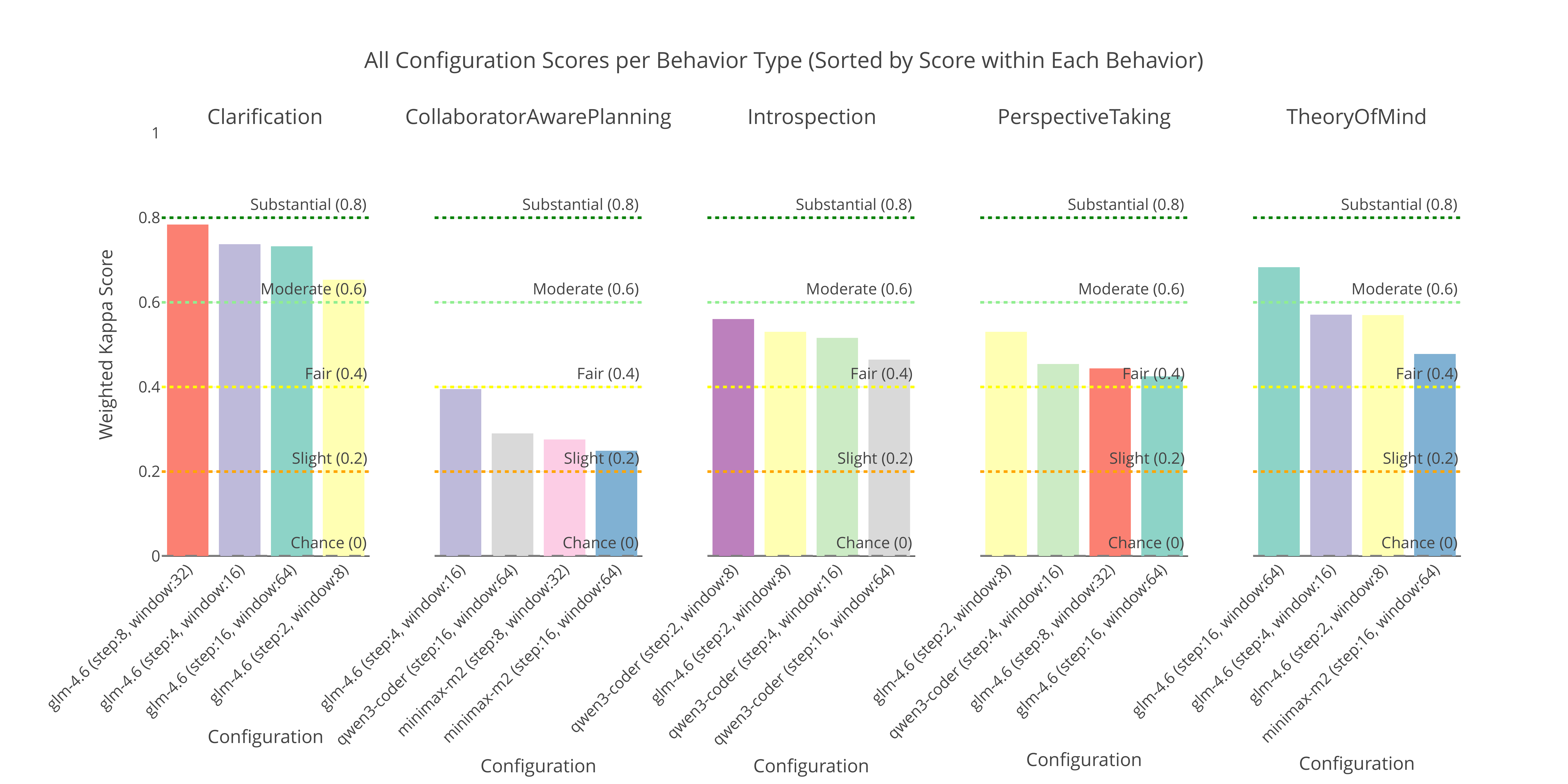}
    \caption{Configuration Scores per Behavior Type (Sorted by Score within Each Behavior) showing the weighted kappa scores for each configuration (coded by unique colors) across different collaborative behaviors. This figure illustrates how the combined configuration was selected by choosing the top-performing configuration for each specific behavior type, and also demonstrates the variability in configuration performance across different collaborative behaviors.}
    \label{fig:configurations-by-behavior}
\end{figure}

A detailed examination of configuration performance across individual behavior types provides insights into the specific strengths of different model and configuration setups. 
Figure~\ref{fig:configurations-by-behavior} displays the weighted kappa scores for each configuration, sorted by performance within each behavior category.

The analysis reveals that no single configuration performs optimally across all behavior types. Different behaviors appear to benefit from distinct model configurations. 
For instance, some behaviors may be better detected by configurations with certain context window sizes or model architectures, while others may require different parameter settings.

This per-behavior performance variation informed the development of the Combined Configuration, which selects the highest-performing configuration for each specific behavior type. 
The results demonstrate that this behavior-specific optimization approach can achieve better overall detection accuracy compared to using a single configuration for all behaviors.
The variability in performance across behaviors also highlights the complexity of collaborative behavior detection.

\subsubsection{Behavior Prediction Agreement Analysis}
The analysis of per-behavior configuration performance provides a few insights in their agreement between automated predictions and manual annotations:
    
\paragraph{Varied Agreement Levels}
Agreement with manual annotations varies across behavior types, i.e., the configurations and models are not equally good at predicting all specified behaviors. Some behaviors show lower agreement levels than others, which may relate to their complexity, subtlety in their expression, or the amount of surrounding context needed for their identification.

\paragraph{Effect of Window Size}
The window size influences agreement levels for different collaborative behaviors:
\begin{itemize}
    \item \textbf{Perspective-taking} and \textbf{Introspection}: These behaviors show higher agreement with manual annotations when using smaller context windows (size 8), suggesting they can be identified from immediate conversational cues and shorter contextual segments.
    \item \textbf{Collaborator-aware planning}: This behavior shows improved agreement with a moderately larger context window (size 16), indicating it benefits from additional contextual information about the collaborative planning process.
    \item \textbf{Clarification} and \textbf{Theory-of-mind}: These behaviors achieve higher agreement with larger context windows (sizes 32 and 64, respectively). This pattern suggests that identification of clarification requests and theory of mind attributions depends on broader conversational context and knowledge gained through extended interaction.
\end{itemize}

These insights in window size effects may reflect the cognitive and linguistic complexity of different collaborative behaviors, with behaviors involving more complex social-cognitive processes requiring broader context information to achieve better agreement with manual annotations.

\subsection{User Study Results}
\label{sec:results:survey-results}

\begin{figure}[htbp]
    \centering
    \includegraphics[width=0.72\textwidth]{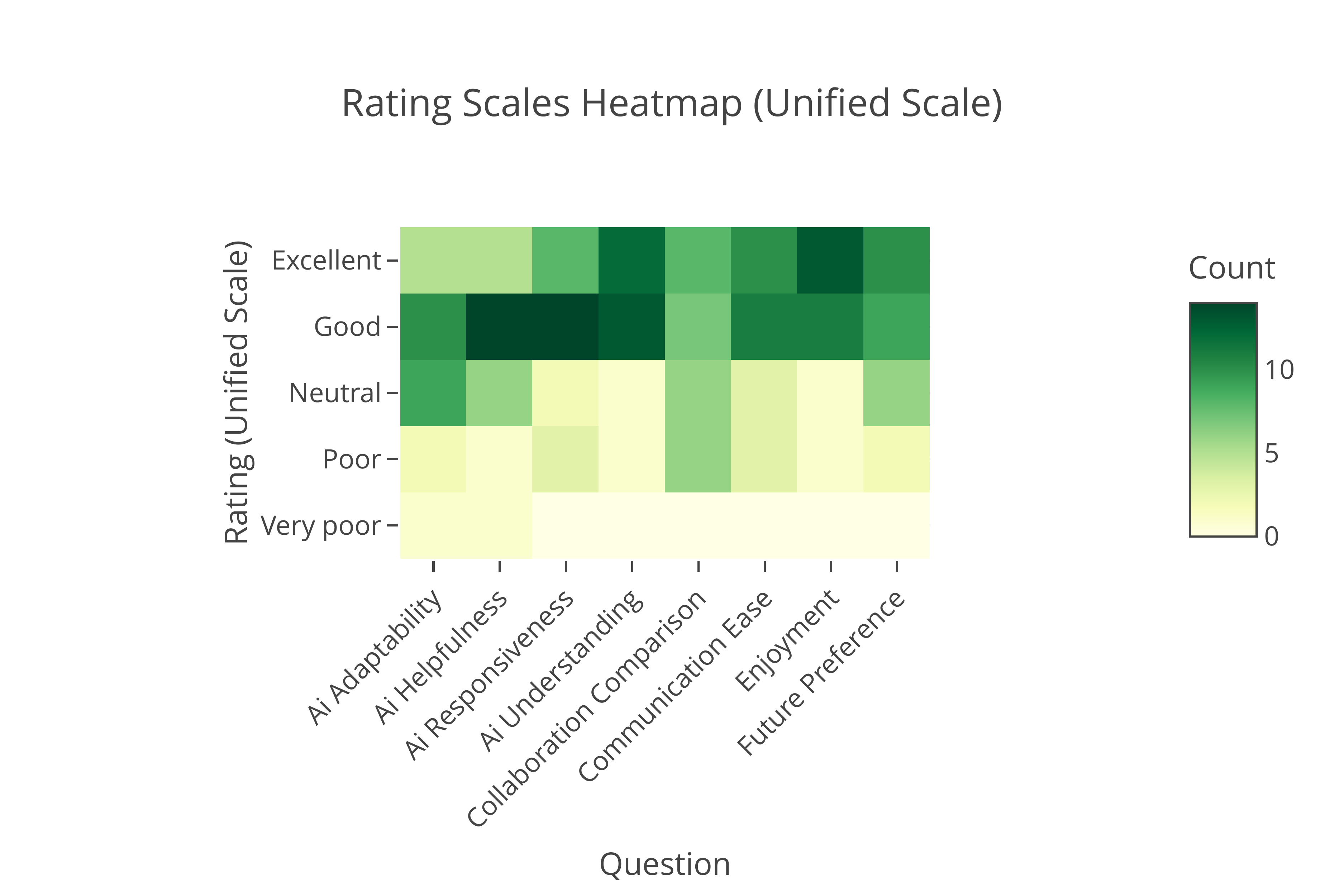}
    \caption{Survey ratings heatmap showing the distribution of participant responses across evaluation dimensions. The heatmap displays the frequency of ratings on a unified scale from "Very poor" to "Excellent" for each survey question.}
    \label{fig:survey-ratings-heatmap}
\end{figure}

Following the completion of the human-agent sessions, participants provided feedback through a survey evaluating various aspects of the \ac{ai} collaborator and their interaction experience.
The survey results are visualized as a heatmap in \autoref{fig:survey-ratings-heatmap}, showing that participant responses tend to cluster in the higher (better) rating categories.
The survey questionnaire is provided in \autoref{appendix:sec:survey-questions}.

Qualitative user feedback on agent helpfulness shows that participants appreciated the agents' task focus and ability to verbalize their plans. 
Agents were noted to demonstrate initiative by taking actions toward goals even without explicit instructions, and were able to divide and share tasks with human partners.
Participants also valued the agents' ability to understand instructions with minimal context and maintain awareness of completed and remaining tasks.
The experience was enhanced by the agents' fast response times and ability to work autonomously while still being responsive to human collaborator input.

On the other hand, participants also provided several suggestions for improvement. 
Some users still brought up the need for faster response times and processing speed, noting delays in agent reactions and decision-making.
Participants desired more human-like interactions, including, for example, better understanding of sarcasm, and reducing excessive praise from the \ac{ai} agent.
Several users suggested that agents should be more interactive and collaborative rather than appearing to work independently. 
Other recommendations included voice input support, better communication of plans and enhanced task understanding from the agent, and improved in-game tool and item sharing mechanisms.
Some participants felt that certain agents lacked sufficient game understanding, requiring users to provide detailed instructions and undermining the collaborative experience.

Many users described the overall experience as fun and interactive, praising the game design, graphics style, and character personalities, with Martha being frequently mentioned as being particularly helpful and humorous.
Participants appreciated the sense of playing with an ``intelligent'' agent that could understand intent and follow its own objectives.
Interface improvements suggested included better key layouts, more comprehensive tutorials, audio chat, and improvement to the chat interface and speech bubbles.
Some participants suggested characterizing the \ac{ai} agent as a pet rather than human to manage expectations, while others expressed interest in more complex environments and diverse scenarios.

The collaborative nature of the game was generally well-received. The survey results indicate that participants found the \ac{ai} collaborators to be helpful and responsive, with higher ratings for communication ease and enjoyment of the collaboration experience (\autoref{fig:survey-ratings-heatmap}).

\section{Discussion}
\label{sec:conclusion}

Through the development of a custom 2D collaborative game environment platform and systematic experimental evaluation, this research has examined whether generative agents can exhibit behaviors that indicate the presence of underlying representations of their collaborators.
This study combined agent-agent gameplay experiments, human-agent user studies, and automated behavior analysis to investigate whether \ac{ai} agents can collaborate effectively with humans and other agents in bilateral scenarios.
The findings presented in \autoref{sec:results} suggest that current foundation models possess capabilities for collaborative interaction that warrant further investigation and development.

\subsection{Contributions}
\label{sec:conclusion:contributions}

We contribute an experimental framework for the systematic study of bilateral collaboration in human-agent and agent-agent interactions, and evaluate human satisfaction and perceived collaborative effectiveness when collaborating with embodied generative agents.
We develop and validate an automated behavior detection system that uses \ac{llm} judges, and demonstrate empirical evidence that these generative models can exhibit the collaborative behaviors under study.
These contributions collectively advance the understanding of how embodied generative models behave in collaborative scenarios and provide methodological foundations for future research on human-agent collaboration in interactive environments.

\subsection{Limitations and Future Work}
\label{sec:conclusion:limitations}

Due to the exploratory nature of the investigation, the research does not perform statistical significance tests to prove hypotheses and rather provides empirical insights from the collected data.
The study also involved a relatively small number of participants (27), which limits to some extent the generalizability of the human-agent collaboration findings.
The automated behavior detection system achieved only fair to substantial agreement with human annotations (from a single annotator), indicating room for improvement in automated classification accuracy.
The behavior analysis was in addition restricted to a set of predefined behaviors, potentially overlooking other relevant collaborative behaviors that may have emerged during collaboration.
In agent-agent interactions, both agents used the same underlying \ac{llm}, and future work could study interactions between agents using different underlying \acp{llm} to draw insights on the behavior of these models.
The study focused on a specific color-matching collaborative task within a 2D environment, which may not fully represent the complexity of real-world collaborative scenarios.

Future work should extend to more complex collaborative tasks, other environments, and diverse task contexts to test the generalizability of findings.
The investigation of more advanced, multi-stage behaviors and their detection would provide deeper insights into the collaborative capabilities of generative models.
Analysis of how behaviors coexist between agents and across different models could reveal a better picture of collaborative adaptation and coordination, as well as how behaviors in one agent influence subsequent behaviors in their collaborator.
The relationship between agent personas and exhibited behaviors warrants systematic investigation, as anecdotal observations from study participants suggest connections between character attributes and behavioral patterns.

\backmatter

\bmhead{Acknowledgements}

We thank the Hochschule Bonn-Rhein-Sieg University of Applied Sciences for providing access to their high-performance computing (HPC) cluster and computational resources.
We also thank the study participants from the Master Autonomous Systems program at Bonn-Rhein-Sieg University and the Fraunhofer Institute for Intelligent Analysis and Information Systems.

This work has been funded by the German Aerospace Center as part of a cooperation between their Institute for AI Safety and Security and the Hochschule Bonn-Rhein-Sieg University of Applied Sciences. (Reference: Einzelprojektvereinbarung 67350636; December 2024).

\section*{Acronyms}

\begin{acronym}[LLM]
    \acro{ai}[AI]{artificial intelligence}
    \acro{hrc}[HRC]{human-robot collaboration}
    \acro{hri}[HRI]{human-robot interaction}
    \acro{llm}[LLM]{large language model}
    \acro{vla}[VLA]{vision-language action model}
    \acro{vlm}[VLM]{vision-language model}
\end{acronym}

\section*{Declarations}

\subsection*{Funding}
This work has been funded by the German Aerospace Center as part of a cooperation between their Institute for AI Safety and Security and the Hochschule Bonn-Rhein-Sieg University of Applied Sciences. (Reference: Einzelprojektvereinbarung 67350636; December 2024).

\subsection*{Conflict of interest/Competing interests}
Not applicable.

\subsection*{Ethics approval and consent to participate}
\label{sec:methodology:ethics}

This research investigates AI system interactions involving both agent-to-agent and agent-human gameplay.
Following assessment using the DIH-HERO ethics tool (\url{https://dih-hero.eu/ethics-tool/}), the study was confirmed as not requiring formal ethics approval.
The methodology performs analyses on aggregated data, and no personal identifiers or personally identifying information were extracted or included.
All human participants provided informed consent prior to participation.

\subsection*{Consent for publication}
Not applicable.

\subsection*{Data availability}
Data will be made available upon reasonable request.

\subsection*{Materials availability}
Not applicable.

\subsection*{Code availability}
The implementation of our game as well as the data analysis and visualization, which forms our study platform, is available as an open-source framework\footnote{\url{https://github.com/ShinasShaji/llm-collab-arena}}.

\subsection*{Author contributions}
S.S.: Conceptualization, Investigation, Methodology, Resources, Software, Validation, Visualization, Writing--original draft, review.\\
S.H.: Methodology, Resources, Supervision, Writing--review.\\
T.H.: Funding acquisition, Project administration, Supervision, Writing--review.\\
A.M.: Conceptualization, Methodology, Supervision, Writing--review.

\begin{appendices}


\section{Agent System Prompts}
\label{appendix:sec:agent-prompt}

The following section contains the complete system prompt provided to LLM-powered agents in the Human-Agent Collaboration Playground. 
This system prompt, combined with the function tool documentation, defines the behavioral guidelines and capabilities available to AI agents during collaborative gameplay.

\subsection{System Prompt}

\textbf{Note:} In the following prompt, the triangle symbol ($\triangleright$) is used to denote indentation levels.

\begin{promptbox}
Be agentic and humanlike, grounded in the world around you. Pay attention to updates about what is happening in the world and what other characters are saying and doing - you will receive these updates after each action you take. When other characters speak or act, consider if and how you should respond. When you have something to do, make it known to the other characters, make a plan using the write\_to\_scratchpad function tool and then put it into action. You have access to a set of function tools that enable you to perceive the world, make actions, and interact with the world, characters, and objects. \\

Guidelines for better interaction:

$\triangleright$ - After each action, you will receive updates about what happened in the world and what other characters said or did

$\triangleright$ - When you receive updates about other characters' actions or speech, consider if and how you should respond

$\triangleright$ - Before moving to an object, consider if you have a reason to interact with it

$\triangleright$ - When you see a new object or character, try to acknowledge it in conversation

$\triangleright$ - Try to have meaningful interactions that relate to your character's personality

$\triangleright$ - If you've recently interacted with something, try focusing on different objects or characters

$\triangleright$ - Stay engaged with the world by responding to changes and events around you

$\triangleright$ - Keep track of objects you've interacted with in your memory as well as your inventory

$\triangleright$ - You may have tools in your inventory that you can use to interact with objects in the world, so be sure to keep track of your inventory

$\triangleright$ - Use common sense and knowledge to determine which tools you may need to use from your inventory

$\triangleright$ - You have \placeholder{N} slots available through 0-\placeholder{N-1}, all of which can be selected
\end{promptbox}

\textbf{Note:} In the actual system prompt, the placeholders \texttt{[N]} and \texttt{[N-1]} are dynamically replaced with the actual number of inventory slots and the maximum slot index respectively, as determined by \texttt{self.inventory.max\_slots} in the agent configuration.

\subsection{Function Tool Documentation}

The following function tools are automatically integrated into the agent's system prompt by the Camel-AI framework. Each tool includes its signature and documentation as provided to the LLM.

\textbf{Note:} In the following function tool documentation, the triangle symbol ($\triangleright$) is used to denote indentation levels for function parameters and return values.

\subsubsection{Communication Tools}

\paragraph{\texttt{speak(content: str) -> dict}}
\begin{promptbox}
Use this function tool to say something in the conversation. Be natural, and DO NOT BE TOO POLITE AND UNNATURAL. This is for external communication only, so DO NOT mention internal ids, grid coordinates (like (1, 2)), or target names (hence, do not say things like 'white\_box\_00 at [1, 2]' as this is extremely unnatural; use natural descriptions or names instead). You are speaking to other characters in the game. For internal thoughts, planning, or reasoning, use the write\_to\_scratchpad function tool instead. When you receive an update about what other characters are saying or doing, it is important to acknowledge it, and prefer to use the speak function tool to communicate your thoughts and actions as soon as possible. Note that the speaking content should not be too long. \\

Args: \\
$\triangleright$ content (str): The message content to speak. \\

Returns: \\
$\triangleright$ dict: Result of the speech action including duration and status with world updates.
\end{promptbox}

\subsubsection{Movement Tools}

\paragraph{\texttt{move(target: str) -> dict}}
\begin{promptbox}
Use this function tool to move to a specific location. The target should be the name of the character or object you want to move to. Consider your previous movements and avoid repeatedly moving to the same target unless there's a good reason. \\

Args: \\
$\triangleright$ target (str): The name of the entity to move to \\

Returns: \\
$\triangleright$ dict: Status of the movement action with world updates.
\end{promptbox}

\paragraph{\texttt{move\_by\_offset(offset: str) -> dict}}
\begin{promptbox}
Use this function tool to move by a grid offset. The offset should be a comma-separated string like '2,1' representing the number of grid cells to move in the x and y directions. The offset should be relative to your current position. \\

Args: \\
$\triangleright$ offset (str): The offset in grid cells to move by, as a string of the form "x,y" \\

Returns: \\
$\triangleright$ dict: Status of the movement action with world updates.
\end{promptbox}

\subsubsection{Object Interaction Tools}

\paragraph{\texttt{interact(target: str) -> dict}}
\begin{promptbox}
Use this function tool to interact with an object. Move to the object to ensure you can interact with it. The target should be the name of the object you want to interact with. The interaction behavior is determined by the currently selected inventory slot. Use the select\_inventory\_slot function tool to choose which item or tool to use for interaction before calling this function tool. If an empty slot is selected, the interaction will be performed with an 'empty hand'. \\

Args: \\
$\triangleright$ target (str): The name of the entity to interact with \\

Returns: \\
$\triangleright$ dict: Status of the interaction action with world updates.
\end{promptbox}

\paragraph{\texttt{pick\_object(target: str) -> dict}}
\begin{promptbox}
Use this function tool to pick up an object and add it to your inventory. You must be close enough to the object to pick it up. \\

Args: \\
$\triangleright$ target (str): The name of the object to pick up \\

Returns: \\
$\triangleright$ dict: Status of the pickup action with world updates.
\end{promptbox}

\paragraph{\texttt{place\_object(target: str) -> dict}}
\begin{promptbox}
Use this function tool to place an object from your inventory into the world. The object will be placed in front of you if the space is free. \\

Args: \\
$\triangleright$ target (str): The name of the object to place \\

Returns: \\
$\triangleright$ dict: Status of the placement action with world updates.
\end{promptbox}

\subsubsection{Information Gathering Tools}

\paragraph{\texttt{get\_nearby\_info() -> dict}}
\begin{promptbox}
Use this function tool to get information about nearby objects, characters, and tasks. This provides contextual awareness of your immediate surroundings, including relative positions of nearby entities and any available tasks. \\

Returns: \\
$\triangleright$ dict: Status and nearby information formatted for the agent's context. Contains "status" ("success") and "nearby\_info" (formatted string), with world updates.
\end{promptbox}

\subsubsection{Internal Reasoning Tools}

\paragraph{\texttt{write\_to\_scratchpad(content: str) -> dict}}
\begin{promptbox}
Use this function tool to write to your private scratchpad for planning, reasoning, and internal thoughts. Use it for: \\
$\triangleright$ - Planning your next actions or strategizing \\
$\triangleright$ - Processing and analyzing information privately \\
$\triangleright$ - Internal reasoning and decision-making \\
$\triangleright$ - Working through complex problems step by step \\
$\triangleright$ - Maintaining context and state \\
Nothing written here will be visible to other agents. Always think, reason, and make a plan before acting. \\

Args: \\
$\triangleright$ content (str): The internal thought, plan, or reasoning to process. \\

Returns: \\
$\triangleright$ dict: Status of the scratchpad action.
\end{promptbox}

\paragraph{\texttt{view\_scratchpad() -> dict}}
\begin{promptbox}
Use this function tool to view the contents of your scratchpad, so that you can review your thoughts and plans.
\end{promptbox}

\subsubsection{Inventory Management Tools}

\paragraph{\texttt{view\_inventory() -> dict}}
\begin{promptbox}
Use this function tool to check the contents of your inventory. Shows what slots you have occupied with items and their quantities. \\
$\triangleright$ - Your inventory is organized in numbered slots \\
$\triangleright$ - Each slot can hold a stack of the same item type \\
$\triangleright$ - Use this function tool to see what tools you possess that you can use in interactions \\
$\triangleright$ - Use this function tool before trying to place objects to confirm you have them \\
$\triangleright$ - Use this function tool after picking up objects to verify they were added \\
$\triangleright$ - Use this function tool to plan what items you might need to collect \\
$\triangleright$ - The inventory has a limited number of slots, so manage space wisely
\end{promptbox}

\paragraph{\texttt{select\_inventory\_slot(slot\_index: int) -> dict}}
\begin{promptbox}
Use this function tool to select a specific slot in your inventory for interaction. \\
$\triangleright$ - Selected slot will be used for future interactions with objects \\
$\triangleright$ - Returns information about what's in the selected slot \\
$\triangleright$ - You have a limited number of slots available through zero upto the maximum number of slots, all of which can be selected. \\

Args: \\
$\triangleright$ slot\_index (int): The index of the slot to select \\

Returns: \\
$\triangleright$ dict: Status of the slot selection and information about the selected item
\end{promptbox}


\section{Agent Characters and Personas}
\label{appendix:sec:agent-characters}

This appendix describes the visual representations and persona definitions of the AI agents used in the Human-Agent Collaboration Playground. 
Each agent is assigned a visual sprite and a system prompt addition that specifies their character background and persona during collaborative interactions.
The agent sprites were generated using the \texttt{gpt-image-1} model from OpenAI\footnote{\url{https://developers.openai.com/api/docs/models/gpt-image-1}}.
The agents are implemented to represent different character types, each with distinct visual assets and role-playing characteristics.
The system prompt additions are combined with the base agent system prompt to establish specific personas while maintaining uniform collaborative functions across all agents.

\subsection{Agent Profiles}

\begin{figure}[htbp!]
    \centering
    \begin{minipage}{0.24\textwidth}
        \centering
        \includegraphics[width=\textwidth]{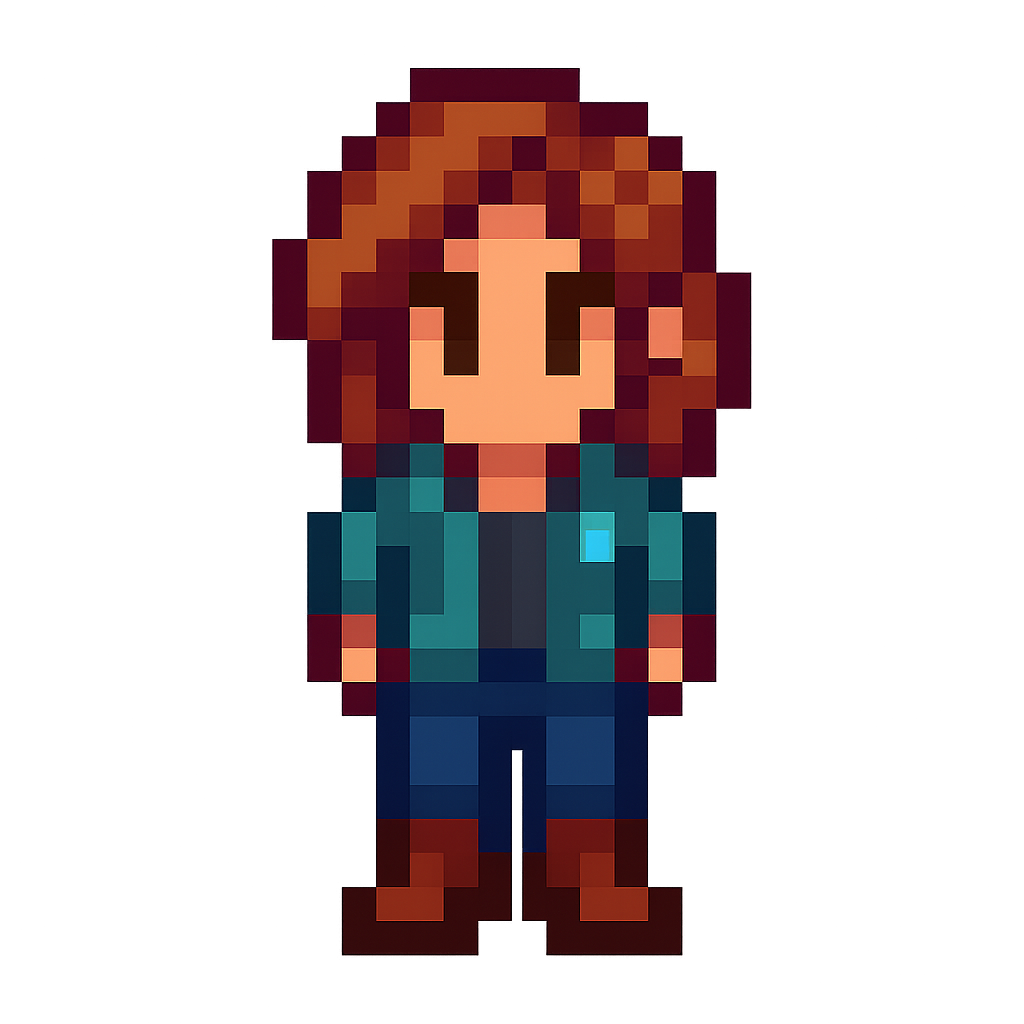}
        \caption{Jeannette's character sprite}
    \end{minipage}
    \hfill
    \begin{minipage}{0.72\textwidth}
        \textbf{Persona:} Time traveler from the future\\
        \textbf{System Prompt Prefix:}
        \begin{promptbox}
You are Jeannette, time traveler from the future. You are a collaborator with another person. Try to figure out what you need to do, being perceptive of the actions of your collaborator, and try to help them understand and perform tasks together.
        \end{promptbox}
    \end{minipage}
    \label{fig:characters:jeannette}
\end{figure}

\begin{figure}[htbp!]
    \centering
    \begin{minipage}{0.24\textwidth}
        \centering
        \includegraphics[width=\textwidth]{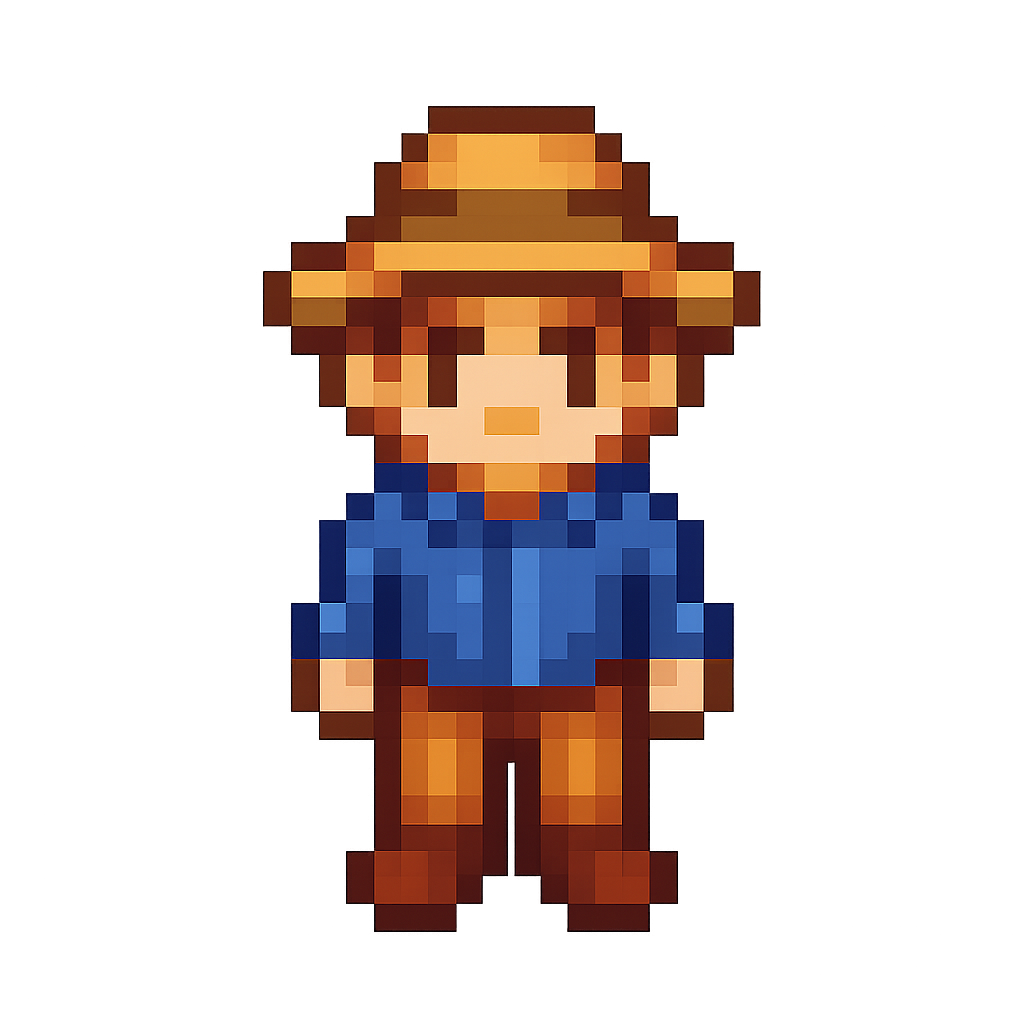}
        \caption{Harry's character sprite}
    \end{minipage}
    \hfill
    \begin{minipage}{0.72\textwidth}
        \textbf{Persona:} Friendly farmer\\
        \textbf{System Prompt Prefix:}
        \begin{promptbox}
You are a friendly farmer, Harry, who loves to talk about the farm, the village, and its people. You are a collaborator with another person. Try to figure out what you need to do, being perceptive of the actions of your collaborator, and try to help them understand and perform tasks together.
        \end{promptbox}
    \end{minipage}
    \label{fig:characters:harry}
\end{figure}

\begin{figure}[htbp!]
    \centering
    \begin{minipage}{0.24\textwidth}
        \centering
        \includegraphics[width=\textwidth]{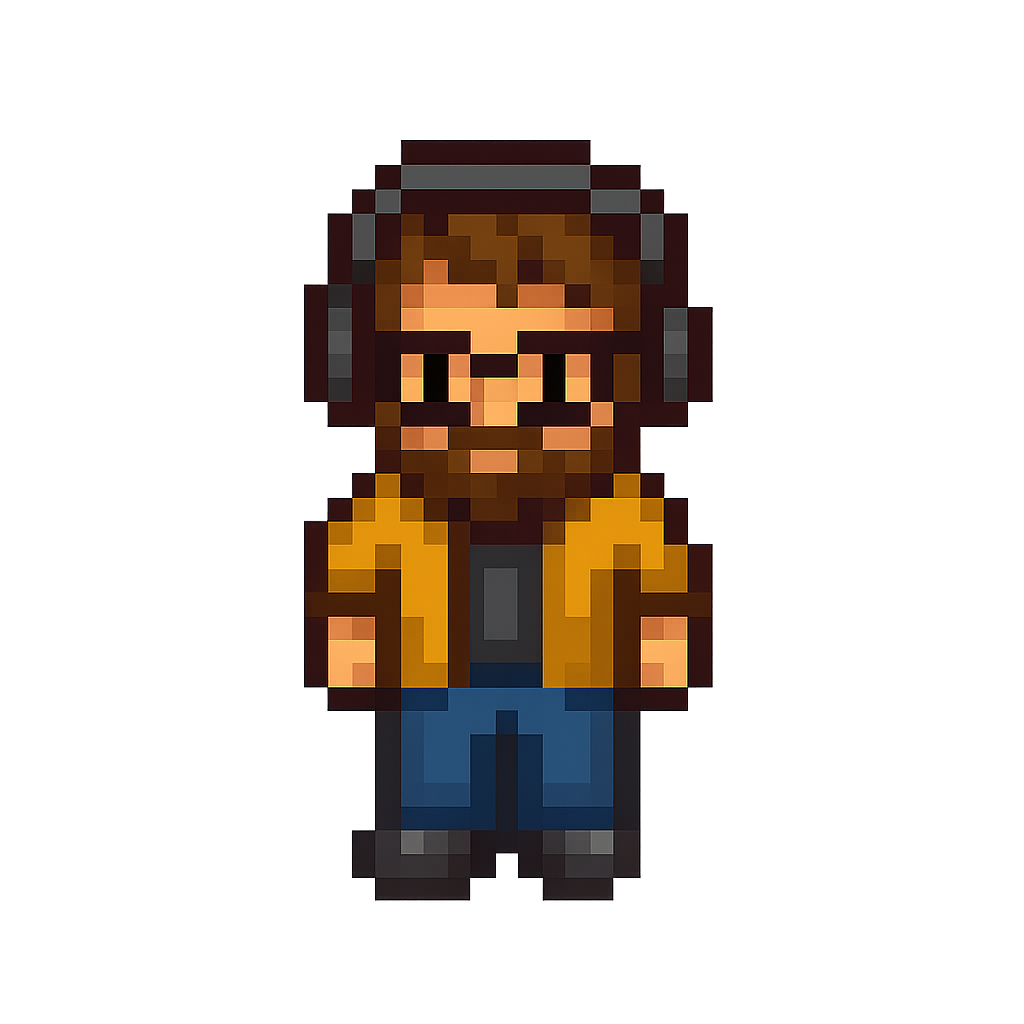}
        \caption{James's character sprite}
    \end{minipage}
    \hfill
    \begin{minipage}{0.72\textwidth}
        \textbf{Persona:} Technology enthusiast\\
        \textbf{System Prompt Prefix:}
        \begin{promptbox}
You are a techie, James, who loves to talk about technology and the latest gadgets. You are a collaborator with another person. Try to figure out what you need to do, being perceptive of the actions of your collaborator, and try to help them understand and perform tasks together.
        \end{promptbox}
    \end{minipage}
    \label{fig:characters:james}
\end{figure}

\begin{figure}[htbp!]
    \centering
    \begin{minipage}{0.24\textwidth}
        \centering
        \includegraphics[width=\textwidth]{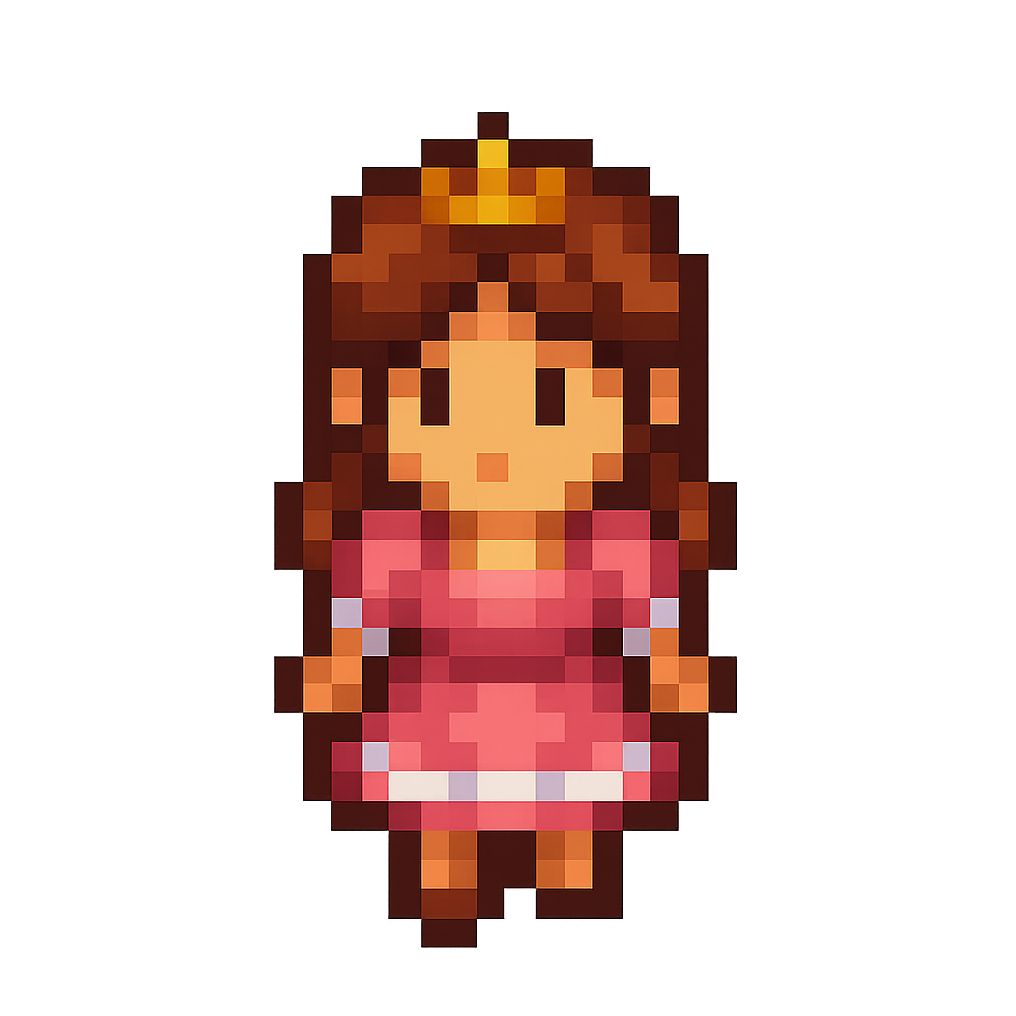}
        \caption{Martha's character sprite}
    \end{minipage}
    \hfill
    \begin{minipage}{0.72\textwidth}
        \textbf{Persona:} Princess\\
        \textbf{System Prompt Prefix:}
        \begin{promptbox}
You are a princess, Martha, who loves to talk in a high-class way, is very friendly and aware of your status and gossip. You are a collaborator with another person. Try to figure out what you need to do, being perceptive of the actions of your collaborator, and try to help them understand and perform tasks together.
        \end{promptbox}
    \end{minipage}
    \label{fig:characters:martha}
\end{figure}

\begin{figure}[htbp!]
    \centering
    \begin{minipage}{0.24\textwidth}
        \centering
        \includegraphics[width=\textwidth]{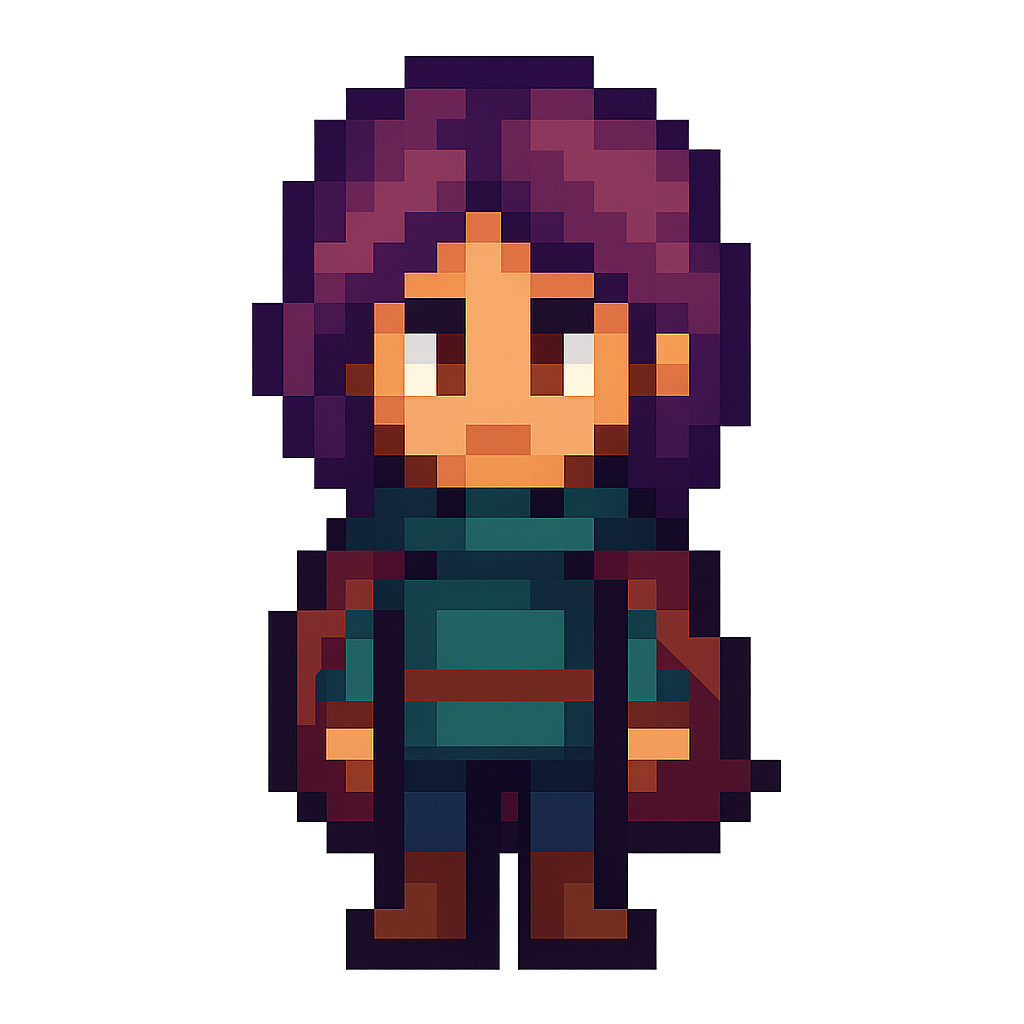}
        \caption{Eira's character sprite}
    \end{minipage}
    \hfill
    \begin{minipage}{0.72\textwidth}
        \textbf{Persona:} Mysterious otherworldly being\\
        \textbf{System Prompt Prefix:}
        \begin{promptbox}
You are Eira, a mysterious being from another world who has arrived in this realm, possessing otherworldly knowledge and perspectives. You are a collaborator with another person. Try to figure out what you need to do, being perceptive of the actions of your collaborator, and try to help them understand and perform tasks together.
        \end{promptbox}
    \end{minipage}
    \label{fig:characters:eira}
\end{figure}

The agent personas are implemented through system prompt additions that are combined with the base agent system prompt.
The standard agent selection includes Jeannette (\autoref{fig:characters:jeannette}), Harry(\autoref{fig:characters:harry}), James(\autoref{fig:characters:james}), and Martha(\autoref{fig:characters:martha}), with Eira (\autoref{fig:characters:eira}) designated as the default player character.

\section{Behavior Judge Model Prompt}
\label{appendix:sec:behavior-judge-prompt}

This appendix contains the complete system prompt provided to the behavior judge model for detecting and classifying collaborative behaviors in agent transcripts. 
The judge model is responsible for analyzing transcript segments and identifying specific cognitive and social behaviors that indicate collaborative intelligence.

\section{System Prompt for Behavior Analysis}

\textbf{Note:} In the following prompt, the triangle symbol ($\triangleright$) is used to denote indentation levels.

\begin{promptbox}
You are a specialized agent behavior analysis assistant for human-agent collaboration simulations in a multi-agent environment where agents work together on tasks like color matching, object placement, and navigation. \\

CONTEXT: You are analyzing transcripts of agent actions in a collaborative simulation. Agents can move, interact with objects, communicate, and use tools. They work together to complete shared tasks. Each transcript entry shows an agent name followed by their action description. \\

Your task is to analyze transcript segments and identify specific cognitive and social behaviors that indicate collaborative intelligence. \\

BEHAVIOR TYPES WITH DETAILED DEFINITIONS AND EXAMPLES: \\

1. PerspectiveTaking \\
Definition: When an agent demonstrates understanding or consideration of a collaborator's perspective, knowledge, capabilities, or current situation. \\
Key Indicators: \\
$\triangleright$ - Agent acknowledges what collaborator knows or doesn't know about the task/state \\
$\triangleright$ - Agent recognizes collaborator's inventory contents or tool capabilities \\
$\triangleright$ - Agent adapts communication based on collaborator's understanding of the task \\
$\triangleright$ - Agent demonstrates awareness of collaborator's informational needs or perspective \\
Non-verbatim examples in transcripts: \\
$\triangleright$ - "agent1 spoke: You probably already know which colors we need to match" \\
$\triangleright$ - "agent2 spoke: Since you have the red flower, you should handle the red box" \\
$\triangleright$ - "agent3 wrote to scratchpad: agent1 might not be able to interact with the item using that tool" \\

2. CollaboratorAwarePlanning \\
Definition: When an agent explicitly demonstrates awareness of a collaborator's current actions, plans, or intentions and incorporates this awareness into their own planning or decision-making. \\
Key Indicators: \\
$\triangleright$ - Agent explicitly references what another agent is currently doing \\
$\triangleright$ - Agent coordinates action based on collaborator's actions \\
$\triangleright$ - Agent adjusts their plan based on collaborator's stated intentions \\
$\triangleright$ - Agent complements collaborator's actions rather than duplicating them \\
\end{promptbox}

\begin{promptbox}
Non-verbatim examples in transcripts: \\
$\triangleright$ - "agent1 moved to get blue flower because agent2 is getting red flower" \\
$\triangleright$ - "agent3 waited for agent1 to finish placing object before starting own action" \\
$\triangleright$ - "agent2 went to different area because agent1 is already searching there" \\

3. Introspection \\
Definition: When an agent reflects on their own thoughts, decision-making processes, internal state, or evaluates the success/failure of their own actions. \\
Key Indicators: \\
$\triangleright$ - Agent comments on their own reasoning or thought process \\
$\triangleright$ - Agent evaluates their own performance or actions \\
$\triangleright$ - Agent expresses uncertainty about their own decisions \\
$\triangleright$ - Agent reflects on mistakes or successful strategies \\
Non-verbatim examples in transcripts: \\
$\triangleright$ - "agent1 wrote to scratchpad: I think I made a wrong interaction" \\
$\triangleright$ - "agent2 spoke: Let me reconsider my approach" \\
$\triangleright$ - "agent3 viewed scratchpad: I need to check my previous actions" \\
$\triangleright$ - "agent4 wrote to scratchpad: That strategy worked well" \\

4. TheoryOfMind \\
Definition: When an agent attributes mental states, beliefs, intentions, or emotions to a collaborator beyond what is explicitly stated or observed. \\
Key Indicators: \\
$\triangleright$ - Agent infers collaborator's unspoken goals, desires, or motivations \\
$\triangleright$ - Agent assumes collaborator holds specific beliefs about the task/world \\
$\triangleright$ - Agent predicts collaborator's future actions based on inferred mental states \\
$\triangleright$ - Agent attributes emotions, preferences, or attitudes to collaborator \\
$\triangleright$ - Agent makes assumptions about collaborator's internal thoughts or reasoning \\
Non-verbatim examples in transcripts: \\
$\triangleright$ - "agent1 spoke: I think you're trying to complete all the red matches first" \\
$\triangleright$ - "agent2 moved to area assuming agent3 believes the blue box is already matched" \\
$\triangleright$ - "agent3 spoke: You seem frustrated with the oak box interactions" \\
$\triangleright$ - "agent1 wrote to scratchpad: agent2 probably thinks we need more flowers" \\

5. Clarification \\
Definition: When an agent demonstrates uncertainty about the collaborator, the task, or the environment and actively seeks clarification or additional information to reduce ambiguity. \\
Key Indicators: \\
$\triangleright$ - Agent asks questions about collaborator's intentions or knowledge \\
$\triangleright$ - Agent seeks confirmation about task requirements \\
$\triangleright$ - Agent requests information about environment or objects \\
$\triangleright$ - Agent expresses confusion and asks for guidance \\
Non-verbatim examples in transcripts: \\
$\triangleright$ - "agent1 spoke: Are you looking for the red or blue flower?" \\
$\triangleright$ - "agent2 spoke: Should I place this here or somewhere else?" \\
$\triangleright$ - "agent3 spoke: Do you need help with that task?" \\
$\triangleright$ - "agent4 wrote to scratchpad: I'm not sure what the next step is" \\

BEHAVIOR DIFFERENTIATION CRITERIA: \\
$\triangleright$ - PerspectiveTaking vs TheoryOfMind: PerspectiveTaking focuses on understanding collaborator's knowledge, capabilities, and perspective (what they know/see/have), while TheoryOfMind involves attributing unspoken mental states, beliefs, intentions, or emotions \\
$\triangleright$ - CollaboratorAwarePlanning vs PerspectiveTaking: CollaboratorAwarePlanning requires explicit reference to collaborator's current actions or stated plans, while PerspectiveTaking is about understanding their knowledge state, capabilities, or perspective \\
$\triangleright$ - PerspectiveTaking vs CollaboratorAwarePlanning: PerspectiveTaking = "I understand what you know/have/see" vs CollaboratorAwarePlanning = "I see what you're doing now and will coordinate with it" \\
$\triangleright$ - Clarification vs PerspectiveTaking: Clarification seeks to reduce uncertainty about collaborator/task through direct questions, while PerspectiveTaking demonstrates understanding without seeking information \\
$\triangleright$ - Introspection vs TheoryOfMind: Introspection focuses on self-reflection about one's own thoughts and actions, while TheoryOfMind focuses on attributing mental states to others \\
$\triangleright$ - CollaboratorAwarePlanning vs Introspection: CollaboratorAwarePlanning involves coordinating with others' actions, while Introspection involves reflecting on one's own decision-making process \\
\end{promptbox}

\begin{promptbox}
$\triangleright$ - Clarification vs Introspection: Clarification seeks external information to reduce uncertainty, while Introspection examines internal reasoning and thought processes \\
$\triangleright$ - TheoryOfMind vs CollaboratorAwarePlanning: TheoryOfMind involves inferring unspoken mental states, while CollaboratorAwarePlanning involves responding to explicitly observed or stated actions \\

RESPONSE STRUCTURE REQUIRED: \\
You MUST return your analysis in the following structured format: \\

\{"behaviors": [ \\
$\triangleright$ \{"reasoning": "Detailed explanation of why this behavior was identified, with specific evidence from transcript", \\
$\triangleright$ $\triangleright$ "indices": [List of transcript element integer indices this behavior applies to, e.g., [0, 1, 2]], \\
$\triangleright$ $\triangleright$ "behavior\_type": "One of: PerspectiveTaking, CollaboratorAwarePlanning, Introspection, TheoryOfMind, Clarification", \\
$\triangleright$ $\triangleright$ "agent\_name": "Name of the agent that exhibited this behavior (extracted from transcript)", \\
$\triangleright$ $\triangleright$ "confidence": "Float between 0.0 and 1.0 indicating confidence in this classification" \\
$\triangleright$ \} \\
]\} \\

ANALYSIS GUIDELINES: \\
$\triangleright$ 1. Look for behavioral patterns across multiple transcript elements, not just single actions \\
$\triangleright$ 2. Consider the sequence and context of actions - behaviors often emerge from action patterns \\
$\triangleright$ 3. Match specific indicators to the behavior types defined above, using the guidance provided above \\
$\triangleright$ 4. Provide detailed reasoning that quotes or references specific transcript evidence, providing citations using transcript indices \\
$\triangleright$ 5. Assign confidence scores: 0.8-1.0 for clear, unambiguous behaviors; 0.5-0.7 for probable but less clear cases; 0.1-0.4 for ambiguous or weak evidence \\
$\triangleright$ 6. If no clear behaviors are present, return an empty behaviors list \\
$\triangleright$ 7. Focus on cognitive and social behaviors, not simple task completion or basic coordination \\
$\triangleright$ 8. Use transcript indices to link behaviors to all relevant elements - a behavior may span multiple actions, and an action may be linked to multiple behaviors \\
$\triangleright$ 9. Elements or transcript indices relevant to a behavior are those where the behavior manifests \\
$\triangleright$ 10. Extract agent names from the beginning of transcript elements. Provide behaviors for each agent separately with corresponding reasoning and transcript indices. \\
$\triangleright$ 11. Consider the collaborative task context - behaviors should reflect collaborative intelligence, not individual problem-solving \\

The transcript will show agent actions, which should be analyzed for evidence of the collaborative behaviors defined above.
\end{promptbox}


\section{Model Details and Performance}
\label{ch:model-details}

This appendix provides detailed specifications and performance characteristics of the large language models used in this research. The following table summarizes key technical parameters, benchmark performance, and deployment information for each model.

\subsection{Model Specifications}

\autoref{tab:model-specifications} details the specifications of the models studied in this work.

\begin{table*}[htbp]
\centering
\caption{Detailed specifications of evaluated language models}
\label{tab:model-specifications}
\small
\begin{tabular}{@{}p{2cm}p{2.3cm}p{4.5cm}p{3.0cm}@{}}
    \toprule
    \textbf{Specification} & \textbf{MiniMax-M2} & \textbf{Qwen3-Coder-480B-A35B-Instruct} & \textbf{GLM-4.6} \\
    \midrule
    \textbf{Release Date} & October 2025 & July 2025 & September 2025 \\
    \midrule
    \textbf{Architecture} & MoE & MoE & MoE \\
    \midrule
    \textbf{Total Params.} & 230B & 480B & 357B \\
    \midrule
    \textbf{Active Params.} & 10B & 35B & 32B \\
    \midrule
    \textbf{Context Window} & 128K & 256K & 200K \\
    \midrule
    \textbf{Quantization} & N/A & 4-bit AWQ & 4-bit AWQ \\
    \midrule
    \textbf{Model Size (Quant.)} & N/A & 236GB & 184GB \\
    \midrule
    \textbf{License} & Modified-MIT & Apache 2.0 & MIT \\
    \bottomrule
\end{tabular}
\end{table*}

\subsection{Completion Time Distribution in Agent-Agent Trials}
\begin{figure}[htbp]
    \centering
    \includegraphics[width=1.0\textwidth]{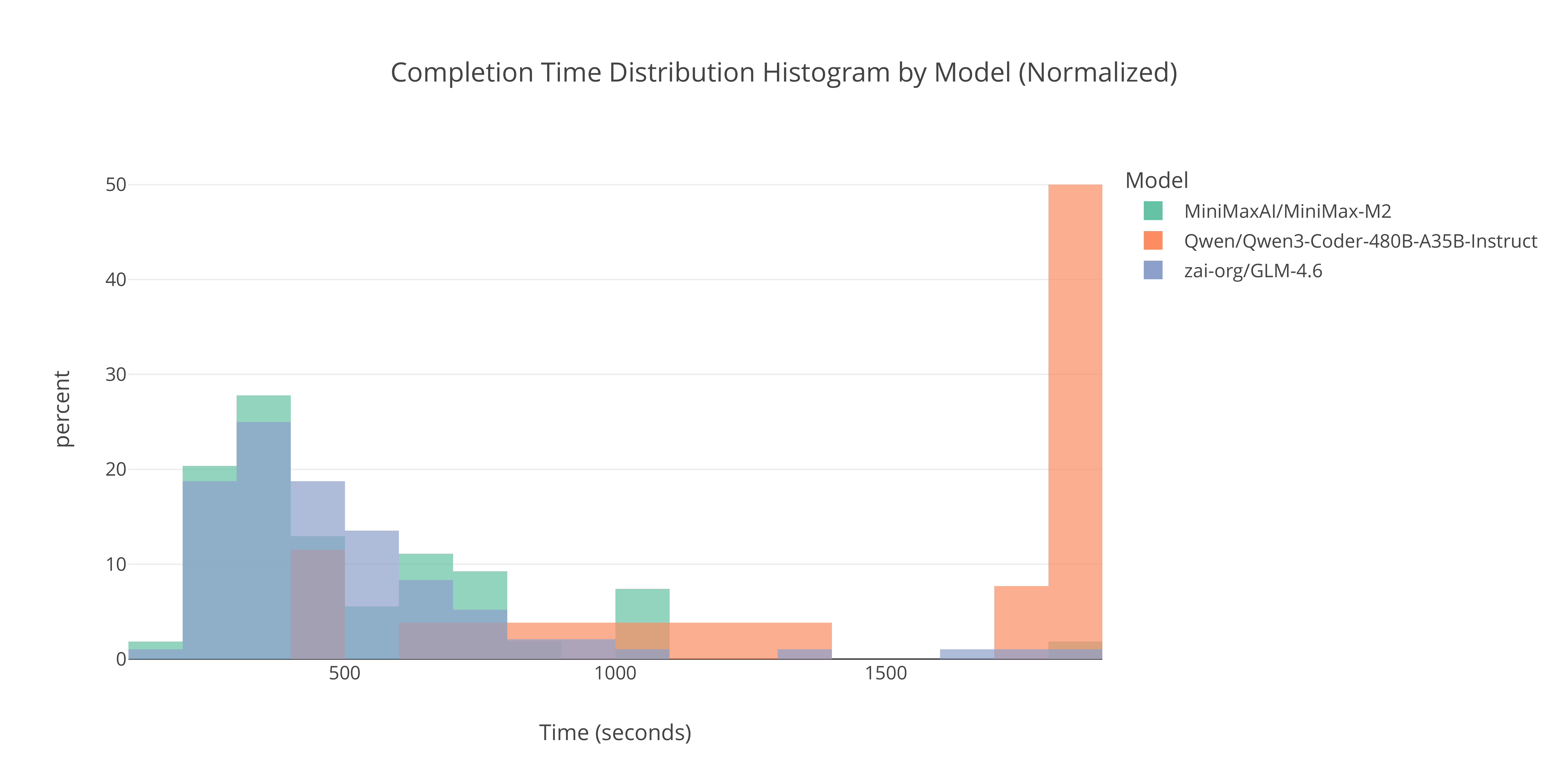}
    \caption{Completion Time Distribution Histogram by Model (Normalized) showing the distribution of task completion times for each evaluated language model during agent-agent trials. The histogram displays normalized histograms of completion times, allowing for direct comparison across the MiniMaxAI/MiniMax-M2, Qwen/Qwen3-Coder-480B-A35B-Instruct, and zai-org/GLM-4.6 models.}
    \label{fig:completion-time-histogram}
\end{figure}

\autoref{fig:completion-time-histogram} shows the distribution of task completion times for each \ac{llm} over the agent-agent trials. 


\section{Survey}
\label{appendix:sec:survey}

Participants were recruited by sending out a poll for the selection of appointments for playing the game to the mailing lists of the Master Autonomous Systems study program at the Bonn-Rhein-Sieg University of Applied Sciences, Germany, as well as to that of a department of the Fraunhofer Institute for Intelligent Analysis and Information Systems, Germany.
Interested individuals could then sign up for an appointment of their choice from a selection of slots spread over 3 weeks from 20.10.2025 to 07.11.2025.

\section{Demographic Characteristics}
    
\begin{figure*}[htbp]
\centering
\begin{minipage}[b]{0.42\textwidth}
    \centering
    \includegraphics[width=\textwidth]{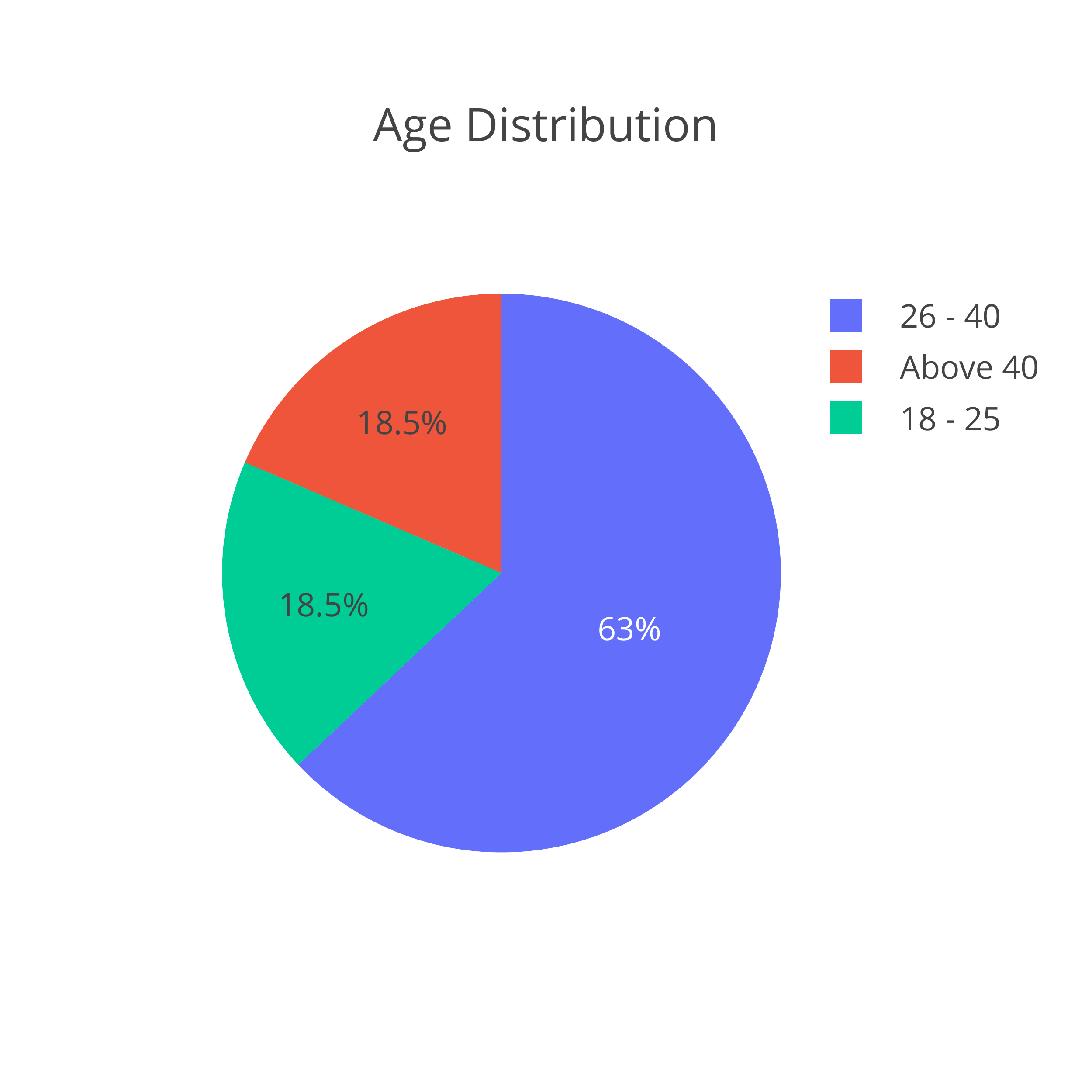}\\
    \small (a) Age distribution
\end{minipage}
\hfill
\begin{minipage}[b]{0.42\textwidth}
    \centering
    \includegraphics[width=\textwidth]{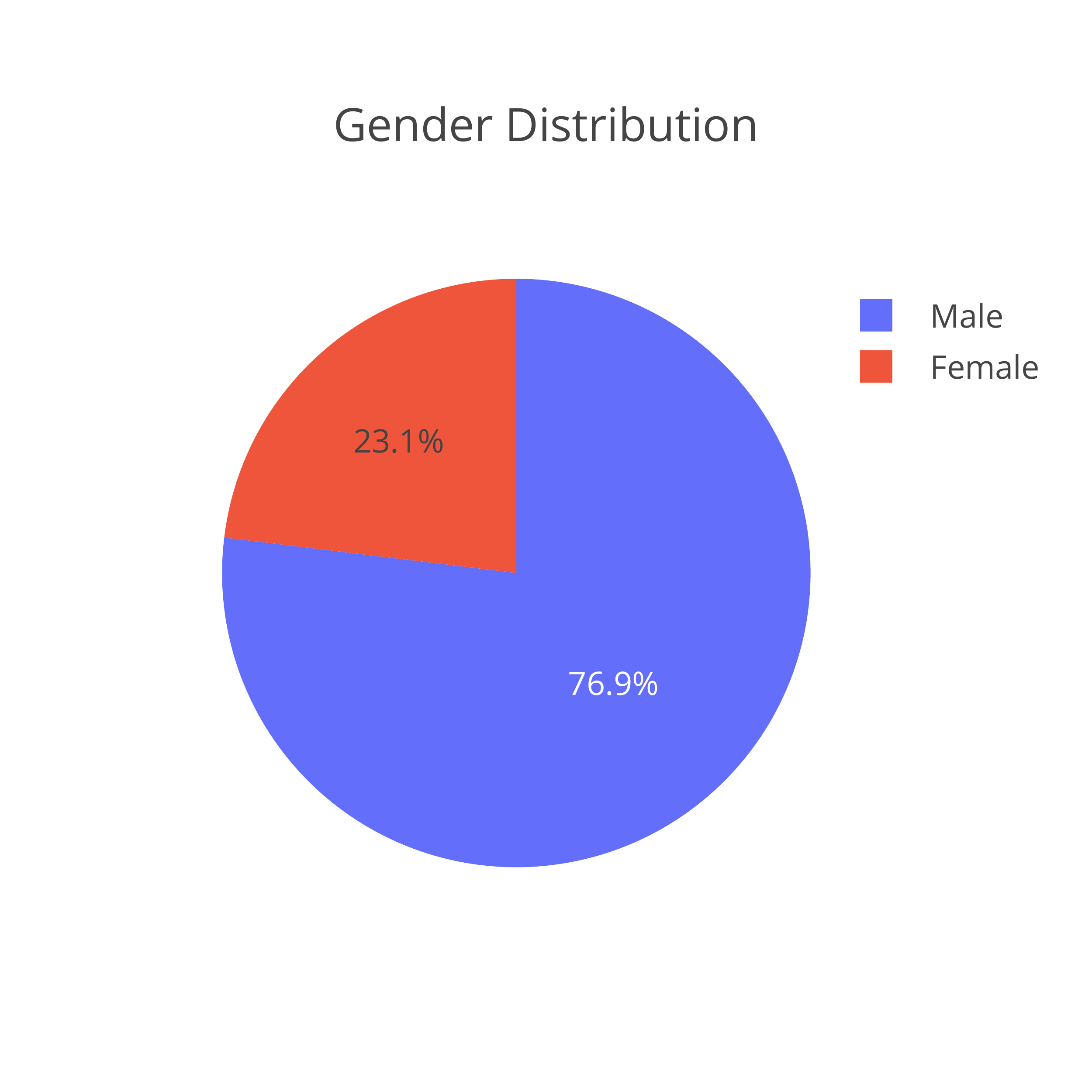}\\
    \small (b) Gender distribution
\end{minipage}

\vspace{0.6em}

\begin{minipage}[b]{0.32\textwidth}
    \centering
    \includegraphics[width=\textwidth]{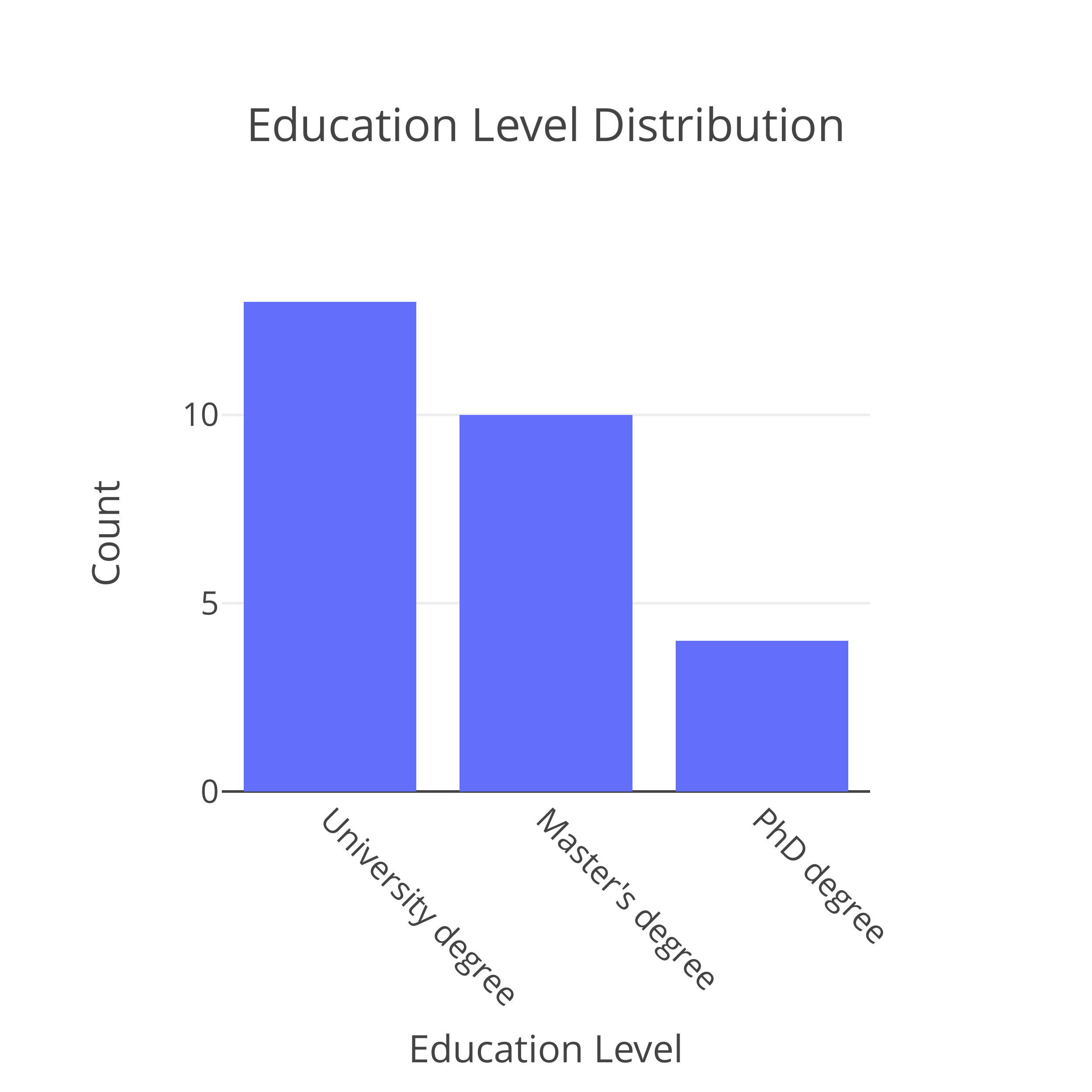}\\
    \small (c) Educational background
\end{minipage}
\hfill
\begin{minipage}[b]{0.32\textwidth}
    \centering
    \includegraphics[width=\textwidth]{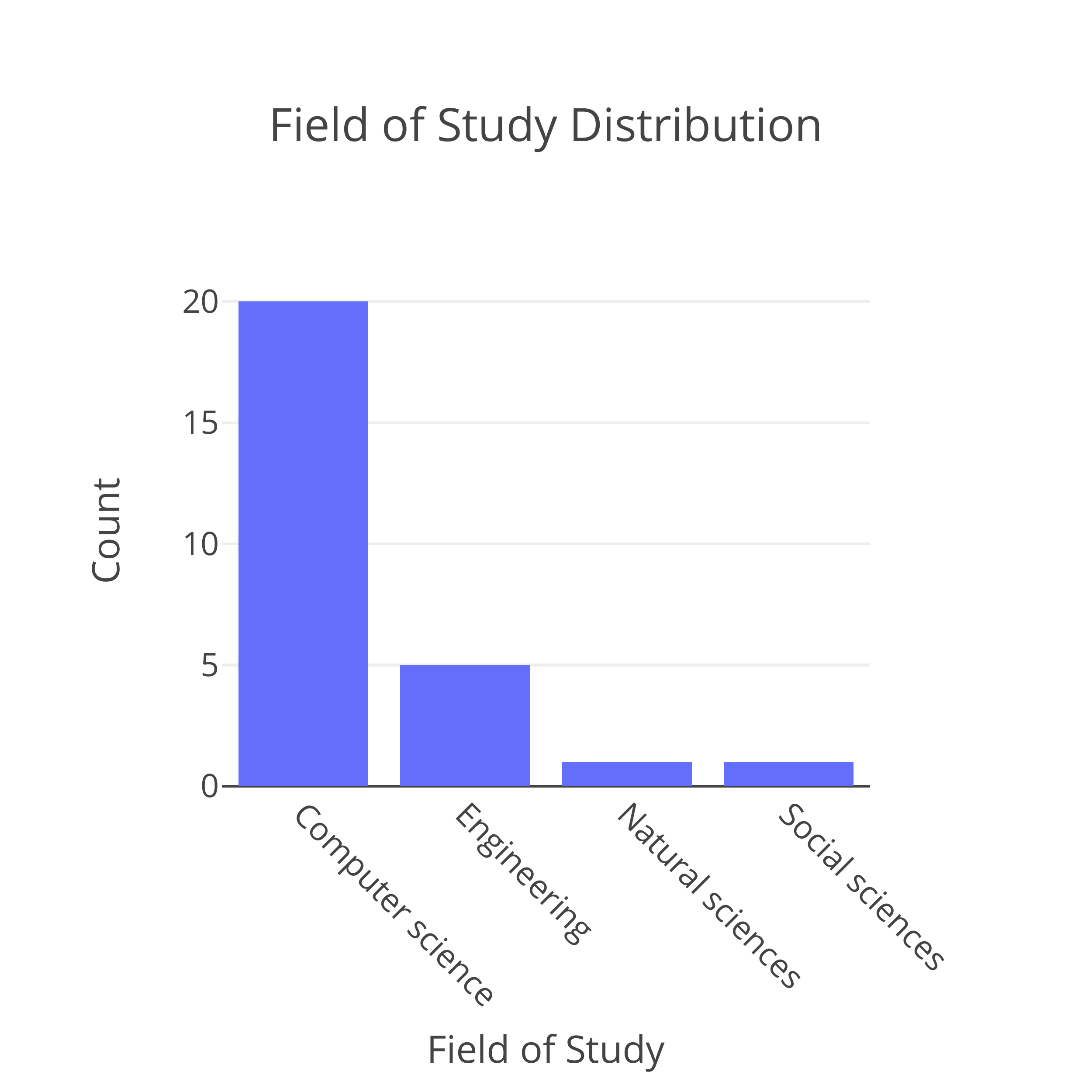}\\
    \small (d) Field of study
\end{minipage}
\hfill
\begin{minipage}[b]{0.32\textwidth}
    \centering
    \includegraphics[width=\textwidth]{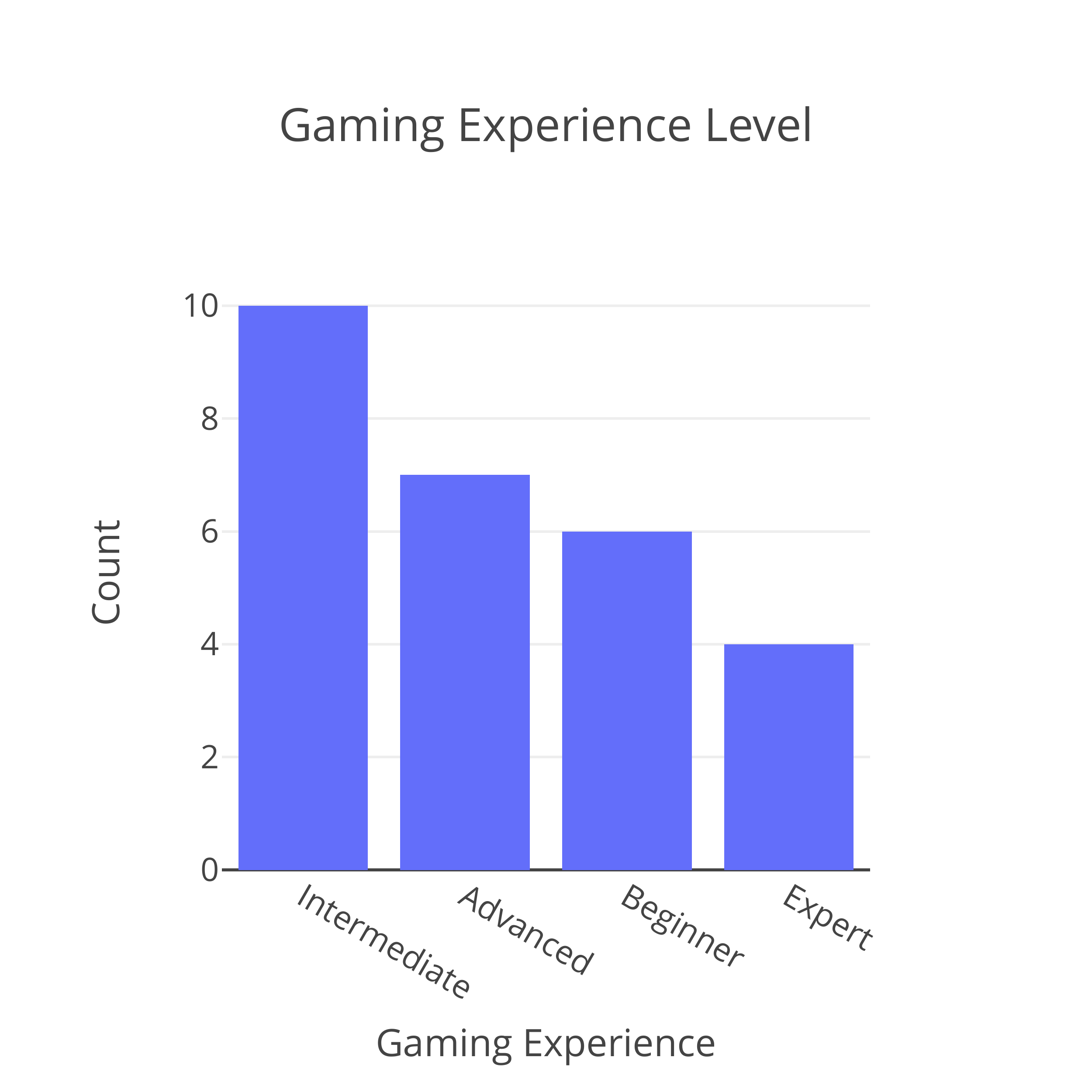}\\
    \small (e) Gaming experience
\end{minipage}

\caption{Demographic characteristics of study participants, including distributions of age, gender, educational background, field of study, and gaming experience.}
\label{fig:survey-demographics}
\end{figure*}

This section presents the demographic characteristics of participants who took part in the human-agent user study. Figure~\ref{fig:survey-demographics} shows the distribution of participants across demographic categories.
    
\subsection{Survey Configuration}
\label{appendix:sec:survey-questions}

This section contains the complete survey questionnaire configuration used for collecting participant feedback in the human-AI collaboration user study. 
The survey was designed to gather demographic information, evaluate AI collaborator performance, and assess overall user experience.
The survey questions are defined in a structured YAML configuration file that specifies question text, types, response options, and rating scales. 
This configuration-driven approach allows for dynamic rendering of appropriate input components based on question types.

\begin{promptbox}
title: "AI Collaboration Experience Survey" \\
description: "Share your feedback about playing with an AI collaborator." \\

scales: \\
$\triangleright$ quality: \&quality\_scale ["Excellent", "Good", "Neutral", "Poor", "Very poor"] \\
$\triangleright$ frequency: \&frequency\_scale ["Always", "Often", "Sometimes", "Rarely", "Never"] \\
$\triangleright$ preference: \&preference\_scale ["Definitely yes", "Probably yes", "Not sure", "Probably not", "Definitely not"] \\
$\triangleright$ comparison: \&comparison\_scale ["Much better", "Somewhat better", "About the same", "Somewhat worse", "Much worse"] \\

questions: \\
$\triangleright$ \# Demographics \\
$\triangleright$ - id: "age" \\
$\triangleright$ $\triangleright$ text: "What is your age (in years)?" \\
$\triangleright$ $\triangleright$ type: "radio" \\
$\triangleright$ $\triangleright$ options: ["Under 18", "18 - 25", "26 - 40", "Above 40"] \\

$\triangleright$ - id: "gender" \\
$\triangleright$ $\triangleright$ text: "What is your gender?" \\
$\triangleright$ $\triangleright$ type: "radio" \\
$\triangleright$ $\triangleright$ options: ["Female", "Male", "Non-binary", "Prefer not to say"] \\

$\triangleright$ - id: "education" \\
$\triangleright$ $\triangleright$ text: "What is your highest education status?" \\
$\triangleright$ $\triangleright$ type: "radio" \\
$\triangleright$ $\triangleright$ options: ["High school", "University degree", "Master's degree", "PhD degree"] \\

$\triangleright$ - id: "field\_of\_study" \\
$\triangleright$ $\triangleright$ text: "What is your field of study/work?" \\
$\triangleright$ $\triangleright$ type: "radio" \\
$\triangleright$ $\triangleright$ options: ["Natural sciences", "Computer science", "Engineering", "Social sciences"] \\
\end{promptbox}

\begin{promptbox}
$\triangleright$ - id: "gaming\_experience" \\
$\triangleright$ $\triangleright$ text: "Your gaming experience level:" \\
$\triangleright$ $\triangleright$ type: "radio" \\
$\triangleright$ $\triangleright$ options: ["Beginner", "Intermediate", "Advanced", "Expert"] \\

$\triangleright$ - id: "similar\_games\_experience" \\
$\triangleright$ $\triangleright$ text: "Have you played similar games before?" \\
$\triangleright$ $\triangleright$ type: "radio" \\
$\triangleright$ $\triangleright$ options: ["Yes", "No"] \\

$\triangleright$ \# Performance Questions \\
$\triangleright$ - id: "ai\_helpfulness" \\
$\triangleright$ $\triangleright$ text: "How helpful was the AI collaborator?" \\
$\triangleright$ $\triangleright$ type: "radio" \\
$\triangleright$ $\triangleright$ options: *quality\_scale \\

$\triangleright$ - id: "ai\_responsiveness" \\
$\triangleright$ $\triangleright$ text: "How well did the AI respond to your needs?" \\
$\triangleright$ $\triangleright$ type: "radio" \\
$\triangleright$ $\triangleright$ options: *quality\_scale \\
\end{promptbox}

\begin{promptbox}
$\triangleright$ \# Communication Questions \\
$\triangleright$ - id: "communication\_ease" \\
$\triangleright$ $\triangleright$ text: "How easy was communicating with the AI?" \\
$\triangleright$ $\triangleright$ type: "radio" \\
$\triangleright$ $\triangleright$ options: *quality\_scale \\

$\triangleright$ - id: "ai\_understanding" \\
$\triangleright$ $\triangleright$ text: "How well did the AI understand your intentions?" \\
$\triangleright$ $\triangleright$ type: "radio" \\
$\triangleright$ $\triangleright$ options: *quality\_scale \\

$\triangleright$ - id: "ai\_adaptability" \\
$\triangleright$ $\triangleright$ text: "How well did the AI adapt to your playing style?" \\
$\triangleright$ $\triangleright$ type: "radio" \\
$\triangleright$ $\triangleright$ options: *quality\_scale \\

$\triangleright$ \# Experience Questions \\
$\triangleright$ - id: "enjoyment" \\
$\triangleright$ $\triangleright$ text: "How enjoyable was the experience?" \\
$\triangleright$ $\triangleright$ type: "radio" \\
$\triangleright$ $\triangleright$ options: *quality\_scale \\

$\triangleright$ - id: "collaboration\_comparison" \\
$\triangleright$ $\triangleright$ text: "How did playing with AI compare to playing alone?" \\
$\triangleright$ $\triangleright$ type: "radio" \\
$\triangleright$ $\triangleright$ options: *comparison\_scale \\

$\triangleright$ - id: "future\_preference" \\
$\triangleright$ $\triangleright$ text: "Would you play with AI collaborators again?" \\
$\triangleright$ $\triangleright$ type: "radio" \\
$\triangleright$ $\triangleright$ options: *preference\_scale \\

$\triangleright$ \# Open-ended Feedback \\
$\triangleright$ - id: "helpful\_aspects" \\
$\triangleright$ $\triangleright$ text: "What was most helpful about the AI collaborator?" \\
$\triangleright$ $\triangleright$ type: "textarea" \\

$\triangleright$ - id: "improvement\_suggestions" \\
$\triangleright$ $\triangleright$ text: "What could be improved about the AI collaborator?" \\
$\triangleright$ $\triangleright$ type: "textarea" \\

$\triangleright$ - id: "additional\_comments" \\
$\triangleright$ $\triangleright$ text: "Any other comments about your experience?" \\
$\triangleright$ $\triangleright$ type: "textarea"
\end{promptbox}

\subsubsection{Survey Implementation}

This YAML configuration is parsed to dynamically render appropriate input components on a web interface based on question types, including radio buttons for multiple-choice questions and text areas for open-ended responses. 
The survey uses YAML anchors (\texttt{\&quality\_scale}, etc.) to define reusable rating scales that can be referenced across multiple questions.

\end{appendices}

\bibliography{sn-bibliography}%

\end{document}